\documentclass[dvipsnames]{article}
\usepackage[a4paper, total={6in, 10in}]{geometry}

\usepackage[numbers]{natbib}

\usepackage[utf8]{inputenc} 
\usepackage[T1]{fontenc}    
\usepackage{hyperref}       
\usepackage{url}            
\usepackage{booktabs}       
\usepackage{amsfonts}       
\usepackage{nicefrac}       
\usepackage{microtype}      

\usepackage{amsmath}
\usepackage{graphicx}
\usepackage{subcaption}
\usepackage{float}
\usepackage{bm}
\usepackage[percent]{overpic} 

\usepackage{xcolor} 
\definecolor{burntorange}{rgb}{0.8, 0.33, 0.0}
\definecolor{carrotorange}{rgb}{0.93, 0.57, 0.13}

\usepackage{multirow}
\usepackage{listings}
\usepackage{enumitem}

\usepackage{bbm}
\usepackage{tabularx}

\usepackage{multicol}

\newcommand{\enc}{\text{enc}}

\usepackage{pifont}
\newcommand{\cmark}{\ding{51}}%
\newcommand{\xmark}{\ding{55}}%

\newcommand{\clevrkiwi}{\emph{CLEVR-MRT}}

\makeatletter
\def\blfootnote{\xdef\@thefnmark{}\@footnotetext}
\makeatother

\title{Visual Question Answering From Another Perspective: CLEVR Mental Rotation Tests}
\author{Christopher Beckham$^{a,b,c}$, Martin Weiss$^{a,c}$, Florian Golemo$^{a,b}$, Sina Honari$^{e}$, \\ Derek Nowrouzezahrai$^{a,d}$ and Christopher Pal$^{a,b,c,\dagger}$}

\begin{document}

\maketitle

\begin{abstract}
Different types of \emph{mental rotation tests} have been used extensively in psychology to understand human visual reasoning and perception. Understanding what an object or visual scene would look like from another viewpoint is a challenging problem that is made even harder if it must be performed from a single image. We explore a controlled setting whereby questions are posed about the properties of a scene if that scene was observed from another viewpoint. To do this we have created a new version of the CLEVR dataset that we call \emph{CLEVR Mental Rotation Tests} (CLEVR-MRT). Using CLEVR-MRT we examine standard methods, show how they fall short, then explore novel neural architectures that involve inferring volumetric representations of a scene. These volumes can be manipulated via camera-conditioned transformations to answer the question. We examine the efficacy of different model variants through rigorous ablations and demonstrate the efficacy of volumetric representations.
\end{abstract}

\section{Introduction}

\blfootnote{$^{a}$Mila - Quebec Artificial Intelligence Institute, $^{b}$ServiceNow Research, $^{c}$Polytechnique Montreal, $^{d}$McGill University, $^{e}$EPFL, $^{\dagger}$Canada CIFAR AI Chair}Psychologists have employed \emph{mental rotation} tests for decades \citep{shepard1971mental} as a powerful tool for devising how the human mind interprets and (internally) manipulates three dimensional representations of the world.  
Instead of using these tests to probe the human capacity for mental 3D manipulation, we are interested here in understanding the ability of modern deep neural architectures to perform mental rotation tasks, and building architectures better suited to 3D inference and understanding. This kind of capability finds application across a variety of visual reasoning and navigation tasks. \blfootnote{Dataset and code will be available at \url{https://github.com/christopher-beckham/clevr-mrt}}

Recent applications of concepts from 3D graphics to deep learning have led to promising results. We are similarly interested in leveraging models of 3D image formation from the graphics and vision communities to augment neural network architectures with inductive biases that improve their ability to reason about the real world. Here we measure the effectiveness of adding such biases, confirming their ability to improve the performance of neural models on mental rotation tasks.  

Concepts from \textit{inverse graphics} can be used to guide the construction of neural architectures designed to perform tasks related to the reverse of the traditional image synthesis processes: namely, taking 2D image input and inferring 3D information about the scene. For instance, 3D reconstruction in computer vision \citep{mvs_tutorial} can be realized with neural-based approaches that output voxel \citep{voxelgan, hologan}, mesh \citep{pixel2mesh}, or point cloud \citep{pointnet} representations of the underlying 3D scene geometry. Such inverse graphics methods range from fully-differentiable graphics pipelines \citep{neural_mesh_renderer} to implicit neural-based approaches with learnable modules designed to mimic the structure of certain components of the forward graphics pipeline \citep{inverse_gfx, deferred_neural}. While inverse rendering is potentially an interesting and useful goal in itself, many computer vision systems could benefit from neural architectures that demonstrate good performance for more targeted mental rotation tasks. 

In our work we are interested in exploring neural ``mental rotation'' by adapting a well known standard benchmark for visual question-and-answering (VQA) through answering questions with respect to another viewpoint. We use the the Compositional Language and Elementary Visual Reasoning (CLEVR) Diagnostic Dataset \citep{clevr} as the starting point for our work.

While we focus on this well known benchmark, many analogous questions of practical interest exist. For example, given the camera viewpoint of a blind person crossing the street, can we infer if each of the drivers of the cars at an intersection can see this person? As humans, we are endowed with the ability to reason about scenes and imagine them from different viewpoints, even if we have only seen them from one perspective. As noted by others, it therefore seems intuitive that we should encourage the same capabilities in deep neural networks \citep{harley2019learning}. In order to answer such questions effectively, some sort of representation encoding 3D information seems necessary to permit inferences to be drawn due to a change in the orientation and position of the viewpoint camera. However, humans clearly do not have access to error signals obtained through re-rendering scenes, but are able to perform such tasks. To explore these problems in a controlled setting, we adapt the original CLEVR setup in which a VQA model is trained to answer different types of questions about a scene consisting of various types and colours of objects. While images from the original dataset are generated through the rendering of randomly generated 3D scenes, the three-dimensional structure of the scene is never fully exploited because the viewpoint camera never changes. We call our problem formulation and data set \emph{CLEVR-MRT}, as it is a new \emph{Mental Rotation Test} version of the CLEVR problem setup. 

\begin{figure}[h]
    \centering
    \begin{subfigure}[b]{\textwidth}
         \centering
         \includegraphics[width=0.31\textwidth]{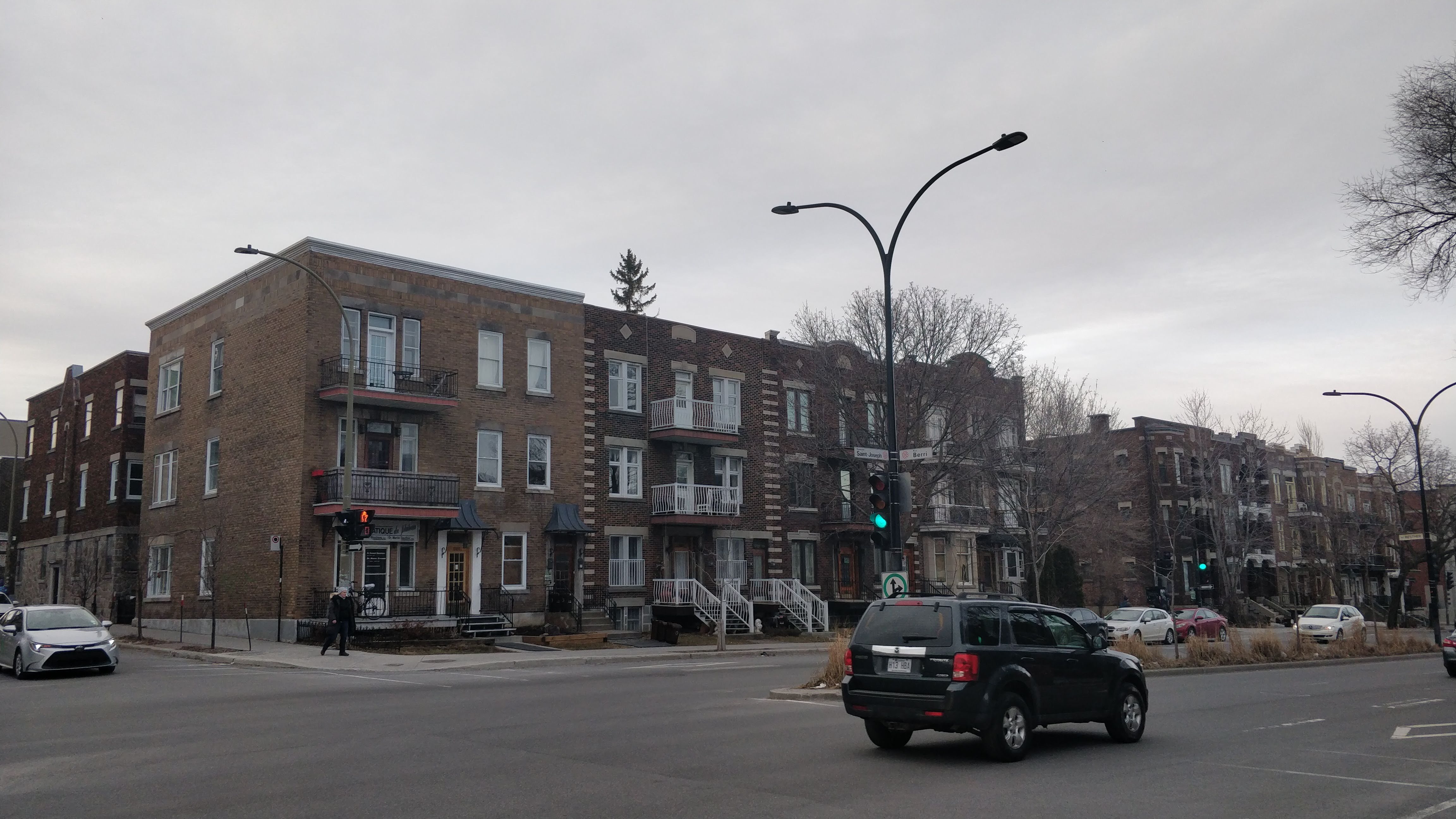}
         \includegraphics[width=0.31\textwidth]{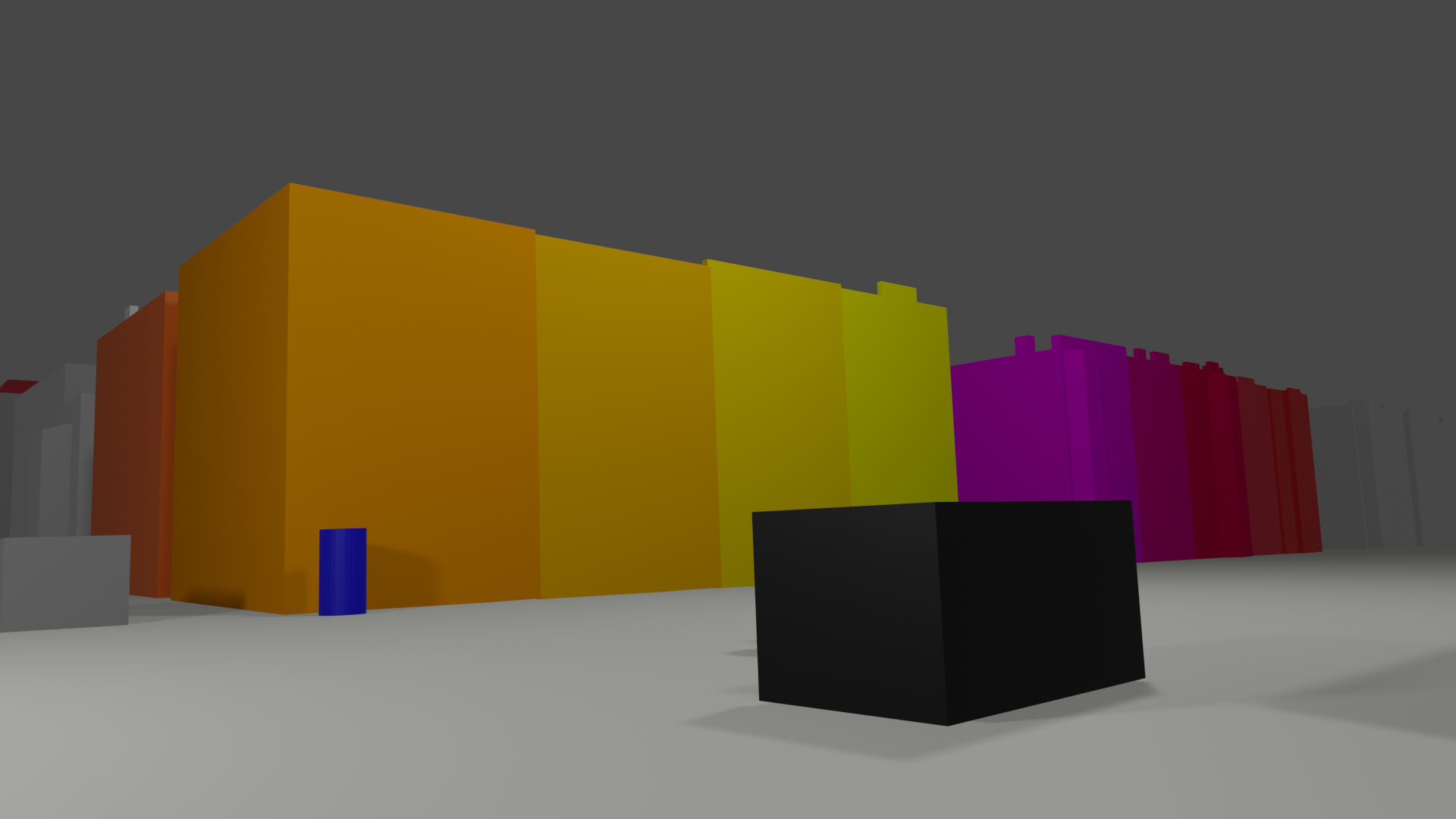}
         \includegraphics[width=0.31\textwidth]{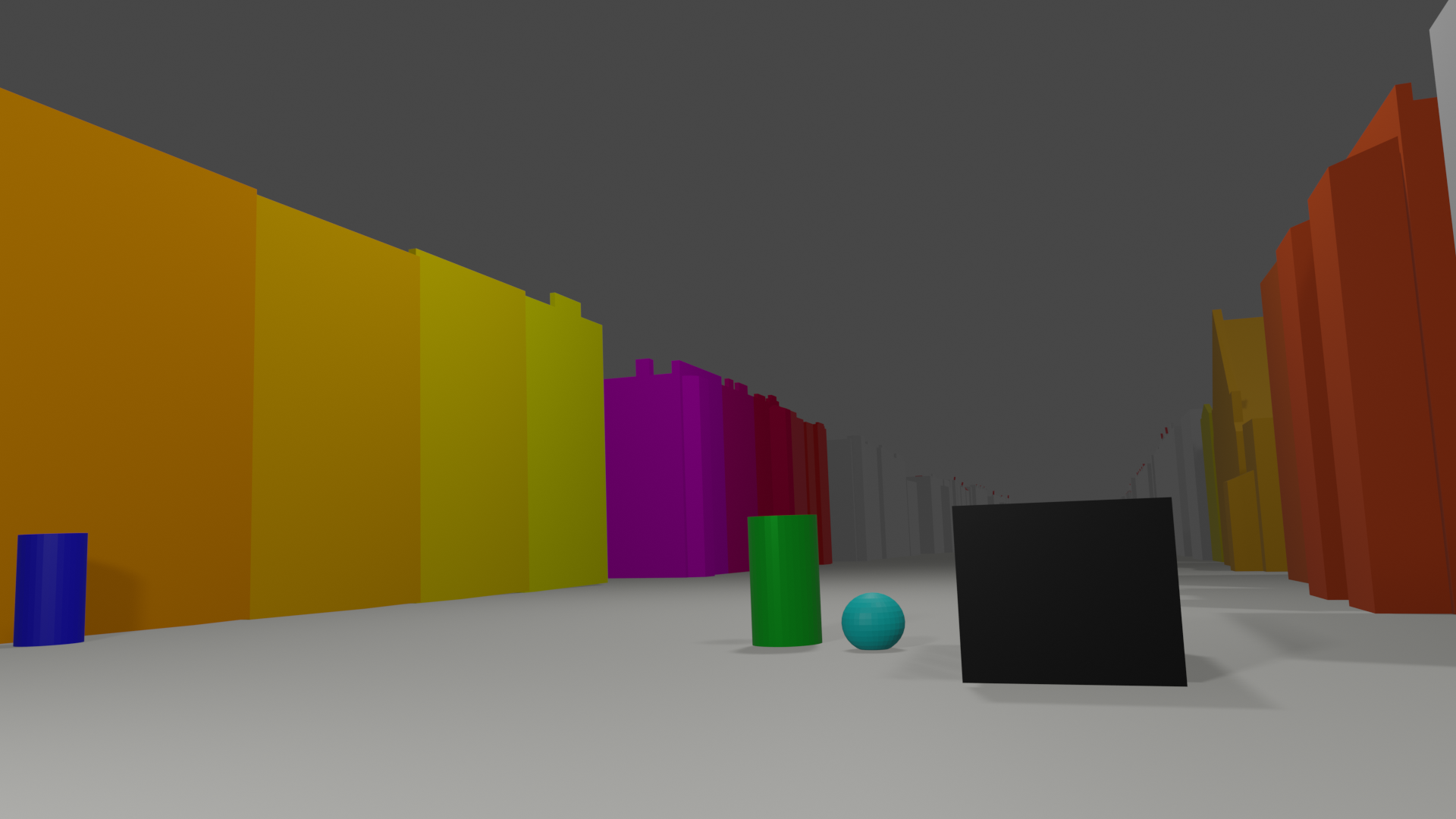}
         \caption{(Left) A view of a street corner. (Middle) a CLEVR-like representation of the scene with abstractions of buildings, cars and pedestrians. (Right) The same virtual scene from another viewpoint, where questions concerning the relative positions of objects after a mental rotation could be of significant practical interest.}
         \label{fig:overview_a}
     \end{subfigure}
    \begin{subfigure}[b]{\textwidth}
        \centering
        \includegraphics[width=0.95\textwidth]{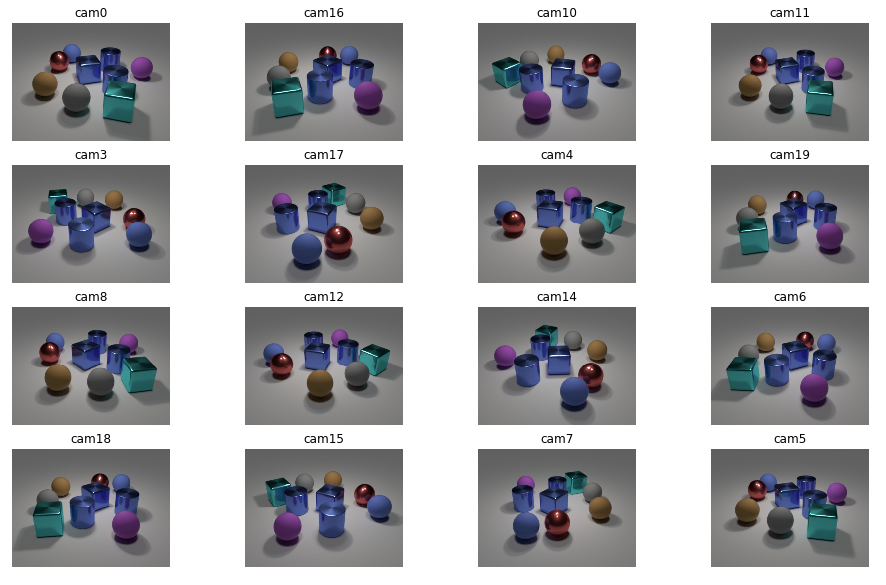}
        \caption{A full example of a CLEVR-MRT scene, showing 16 randomly sampled views of the scene (out of 20 in total). The canonical view is not shown here, but a sample of questions pertaining to the canonical view are: 
        \textbf{Q}: Are there the same number of large gray objects that are left of the gray thing and big matte cylinders? \textbf{(True)};
        \textbf{Q}: Is the number of large brown things that are behind the red thing less than the number of purple objects? \textbf{(True)};
        \textbf{Q}: Are there more blue blocks to the left of the large gray thing than blue matte spheres? \textbf{(False)};
        \textbf{Q}: Are there an equal number of big brown things that are in front of the large brown rubber ball and big metal things that are behind the purple sphere? \textbf{(False)}
        }
         \label{fig:overview_b}
     \end{subfigure}
    \caption{(a) A real-world example where the ability to perform mental rotations can be of practical utility. (b) Images from a randomly selected scene from the \emph{CLEVR-MRT} dataset.}
    \label{fig:overview}
\end{figure}

In CLEVR-MRT alternative views of a scene are rendered and used as the input to a perception pipeline that must then answer a question that is posed with respect to  another (the original CLEVR) viewpoint. This gives rise to a more difficult task where the VQA model must learn how to map from its current viewpoint to the viewpoint that is required to answer the question. In Figure \ref{fig:overview_a}, we motivate our dataset by depicting a real world street corner and a `CLEVR-like' illustration of the scene, where questions concerning the relative positions of objects after a mental rotation could be of practical interest (e.g. intelligent intersections, cars, robots, or navigation assistants for the blind), and in Figure \ref{fig:overview_b} an actual scene from \emph{CLEVR-MRT}. 

Using our new mental rotation task definition and our CLEVR-MRT dataset, we examine a number of new inverse-graphics inspired neural architectures. We examine models that use the FILM (Feature-wise Linear Modulation) technique \citep{film} for VQA, which delivers competitive performance using contemporary state-of-the-art convolutional networks. We observe that such methods fall short for this more challenging MRT VQA setting. This motivates us to create new architectures that involve inferring a latent \emph{feature volume} that we subject to rigid 3D transformations (rotations and translations), in a manner that has been examined in 3D generative modelling techniques such as spatial transformers \citep{stn} as well as HoloGAN \citep{hologan}. This can either be done through the adaptation of a pre-trained 2D encoder network, i.e. an ImageNet-based feature extractor as in Section \ref{sec:proj_2d_to_3d}, or through training our encoder proposed here, which is obtained through the use of contrastive learning as in Section \ref{sec:contrastive}. In the case of the latter model, we leverage the InfoNCE loss \citep{oord2018representation} to minimise the distance between different views of the \emph{same} scene in a learned metric space, and conversely the opposite for views of \emph{different} scenes altogether. However, rather than simply using a stochastic (2D) data augmentation policy to create positive pairs for the contrastive loss (e.g. random crops, resizes, and pixel perturbations), we leverage the fact that we have access to many views of each scene \emph{at training time} and that this can be seen as a data augmentation policy operating in 3D. This in turn can be leveraged to learn an encoder that can map 2D views to a 3D latent space without assuming any extra guidance such as camera extrinsics.

Note that the specific formulation of the VQA task differs slightly from the analogy we have proposed, due to technical and pragmatic reasons: rather than having the viewpoint camera be unknown and the canonical camera being giiven or inferred from a `landmark', we instead have the opposite but slightly less intuitive interpretation, which is that the viewpoint camera is known and the canonical viewpoint is unknown. This is just a minor difference however. As we will see later, due to the many views available per scene and the fact that this task is supervised with respect to question/answer pairs, the problem can still be addressed.

\subsection{Related work} \label{sec:related}

Several extensions of the CLEVR dataset exist, though they mainly focus on extensions to the language processing elements of the problem setup, exploring themes such as: systematic generalisation \citep{closure}, adding dialogue \citep{clevr_dialog}, and robust captioning of changes between scenes \citep{robust_caption}. In terms of visual-based extensions, \cite{clevrer} proposed a temporal version of CLEVR which looks at VQA in the context of causal and counterfactual reasoning.  Concurrent to our work, a version of CLEVR has recently been proposed \citep{clevr_avs} in the context of reinforcement learning, where an agent is trained to perform viewpoint \emph{selection} on a scene to be able to answer the question, with each scene consisting of a large occluder object in the center to accentuate occlusions. However, the main difference is that our dataset decouples the camera viewpoint from the viewpoint from which the question must be answered. Furthermore, their dataset has relatively limited question and scene variability (for instance, focusing on only two types of questions and the same occluding object in the center). We also do not assume the VQA model is an agent that is able actively change its viewpoint to better answer the question -- instead, our model must learn to `imagine' what the same scene should look like from another perspective, conditioned only on a single view. The most closely related work to ours explores the incorporation of 3D information into a FILM-based pipeline \cite{clevr_yue2019}, though this is done by conditioning on multiple views of the same scene at inference time either through pooling the features of those views or through a scene representation network \citep{gqn}. Their main motivating factor for their work is to address the issue of occlusions which cannot be easily resolved under a single view setting. In contrast we examine mental rotation-based reasoning where the input is a single image and a 3D latent volume has to be inferred from it. Lastly, their proposed dataset has limited variability compared to ours, with only four equally spaced camera rotations (every $90^{\circ}$) at a fixed elevation.

Our work is very closely related to single view reconstruction \cite{fahim2021single} because at inference time the VQA model is only being conditioned on a single view, and so the network must infer as much as possible about the scene in order to answer the question. While single view reconstruction constitutes a very difficult learning scenario, the requirement that only a single view be needed makes it a very interesting and pragmatic line of research for problem domains where data collection is difficult.  Single view reconstruction has a wide variety of applications ranging from 3D facial reconstruction \cite{srface_jo2015single, srface_dou2018monocular}, to pose estimation for anatomical structures in medical imagery \cite{srpose_kang2016simultaneous}, to image super-resolution \cite{srres_he2023single, srres_behjati2023single}, and the reconstruction of 3D objects in general \cite{srmesh_yan2016perspective, srmesh_pontes2018image2mesh}. Since single views make it impossible to resolve issues relating to occlusion, prior information must be integrated into the learning algorithm to infer any missing details. Classically this is done through hand-crafted and highly engineered solutions. In the case of deep neural networks however one way this can be achieved is through transfer learning, where a network that is pre-trained on one task is repurposed for another. It is usually assumed that the new task contains relatively fewer examples, labels, or lower quality data than the former, hence the need to `transfer' knowledge to the latter. For instance, \cite{srmesh_pontes2018image2mesh} considers the task of performing 3D reconstruction of an indoor scene from a single 2D image. Their architectures leverages a Mask R-CNN backbone \cite{he2017mask} that was originally trained on MS-COCO ($\ge$ 300K images and rich labels), which is subsequently repurposed for their indoor scene dataset. Since MS-COCO constitutes an ample number of real world occlusions, it is assumed that knowledge about how to resolve them (baked into the pre-trained R-CNN network) can be repurposed for a smaller but more specialised dataset, in this case indoor scenes. Similarly in our work, we consider two types of pre-trained network for our VQA pipeline: an ImageNet classifier, and our own which leverages contrastive learning. 

In \cite{srmesh_pontes2018image2mesh}, they also explore the way in which visual question answering techniques could be used to enhance the quality of 3D scene representations based on traditional computer graphics CAD models. In their work these questions really serve as a form of auxiliary task, aiding their primary goal of creating these CAD based scene representations. In contrast, in our work we focus on learning completely neural representations in which the final goal is always that of answering a question regarding the scene, encoded in natural language. Recent work on Neural Radiance Fields \cite{nerf} has shown great promise for representing pixel level details of 3D scenes, allowing scenes to be rendered from novel viewpoints using fully neural representations. In contrast, our work, focuses exclusively on visual question answering as the final goal and represents a setting where a pixel level model of the scene from an alternative viewpoint is not needed. For problems like high level navigation, e.g. directing a robot or a person to location containing an object at a particular location), our method operating at a completely semantic level of abstraction, allows models of lower complexity to be used because our models do not need to reconstruct new viewpoints.       

The use of rigid transforms to infer latent 3D volumes was loosely inspired by HoloGAN \citep{hologan}. Here, they use a GAN to map a randomly sampled noise vector to a 3D latent volume (a voxel representation) before subsequently rendering using a neural renderer. Several works \citep{deepvoxels, voxel_facebook} condition on images and camera poses to learn voxels that represent the input, with options to re-render from different camera poses. These methods, however, assume a dataset consisting of just a single scene as well as camera poses that are known. Conversely, CLEVR consists of tens of thousands of scenes, which makes any re-rendering task significantly more difficult due to the need for the renderer to generalise to all scenes. Rather than considering an approach that does both encoding and decoding (re-rendering), we only consider encoding, which is more computationally efficient. 3D latent volumes can be seen as highly compressed and feature-rich representations of their original images however, and can in principle be used to re-render a scene \cite{yang2022exploring}.

Other ways of encoding 3D data can be used such as point clouds \citep{pointnet}, meshes \citep{pixel2mesh, neural_mesh_renderer}, surfels \citep{pix2shape}, latent codes \citep{gqn}, or as an implicit neural representation such as in \citep{nerf}. Out of these modalities, representing data as 3D volumes is convenient because they can be used in conjunction with 3D convolutions without modification. 

In terms of VQA, other models have been proposed, e.g., MAC \citep{mac} proposes a memory and attention-based reasoning architecture for more interpretable VQA. While this could in principle be modified to leverage 3D volumes, FILM serves as a simpler architectural choice for analysis. More sophisticated architectural choices can also involve strongly-supervised feature extractors (e.g. Mask R-CNN \citep{nie2020shallow2deep, nsvqa} or bounding box predictors \citep{mdetr}) or neuro-symbolic reasoning pipelines \citep{nsvqa}, but here we opt for a simple FILM-based architecture where the feature extractor is `simple' (one whose training does not involve `rich' labels like segmentation masks or bounding boxes, such as a pre-trained ImageNet classifier). 

As for learning encoders, there has been a lot of interest in leveraging self-supervised learning as a way to learn encoders that are just as competitive as their `supervised' counterparts. In the case of contrastive learning \cite{oord2018representation} -- one particular instance of self-supervised learning -- we do not assume labels for individual images but assume it is possible to define labels with respect to pairs of images. In particular, these labels can either be positive or negative, denoting some semantic relationship between the pair (e.g. does this pair of images belong the same category or not?). The main objective is to learn a latent space in which pairs of inputs that should be positive are close together in that latent space, and conversely the opposite for negative pairs. Once trained, the encoder can be treated as a feature extractor to be used for future downstream tasks. In \cite{amdim} the authors propose that such a framework can be used to maximise the mutual information across different \emph{views} of an input, for instance different camera views within a scene, or different modalities corresponding to the same input (e.g. olfactory, visual). In \cite{amdim, contrastive_hinton} the authors demonstrated that when positive pairs comprise stochastic 2D data augmentation operations on the same image then a resulting classifier trained on that encoder can obtain performance on par with that of its purely supervised counterpart for many benchmark image datasets including ImageNet. In \cite{moco}, competitive results were achieved with respect to object segmentation and detection. One particular bottleneck that is common with these techniques is that of memory, since a large batch size is usually needed in order to contrast each positive pair with a significantly larger number of negative pairs, though this has been mitigated with more recent methods \cite{moco}.

In terms of combining such techniques with 3D, \cite{contrastive_multiview} explored contrastive learning of scenes, though the multi-view aspect in this setting was applied to different sensory views rather than camera views. Lastly, \cite{harley2019learning} explored the use of contrastive learning on 2.5D video (i.e. RGB + depth) to predict novel views, with the goal of learning 3D object detectors in a semi-supervised manner.

\subsection{Proposed dataset}

The CLEVR dataset \citep{clevr} is a VQA dataset consisting of a range of synthetic 3D shapes laid out on a canvas. The dataset consists of a range of questions designed to test various aspects of visual reasoning such as counting (e.g. `how many red cubes are in this scene?'), spatial relationships (e.g. `what colour is the cylinder to the left of the big brown cube?') and comparisons (e.g. `are there an equal number of blue objects as red ones?'). In recent years however, proposed techniques have performed extraordinarily well on the dataset \citep{film, mac}, which has inspired us to explore VQA in more difficult contexts. 

\begin{figure}[h]
    \centering
    \begin{subfigure}[b]{0.6\textwidth}
        \includegraphics[width=\textwidth]{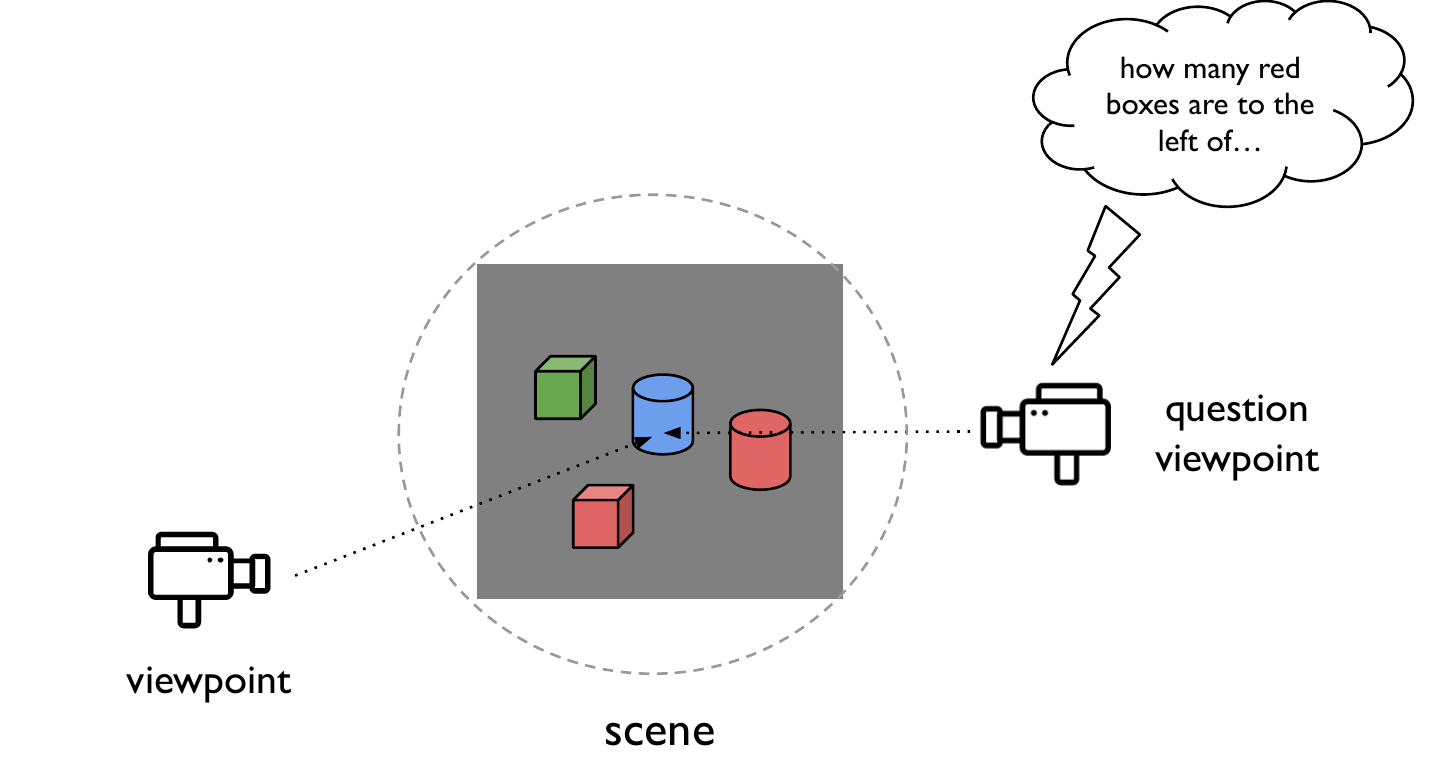}
        \caption{Dataset schematic.}
        \label{fig:dataset_illustration_a}
    \end{subfigure}
    \par\bigskip
    \begin{subfigure}[b]{0.9\textwidth}
        \includegraphics[width=0.5\textwidth]{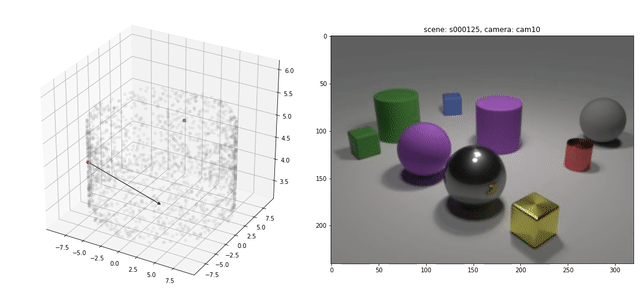} \ \ \includegraphics[width=0.5\textwidth]{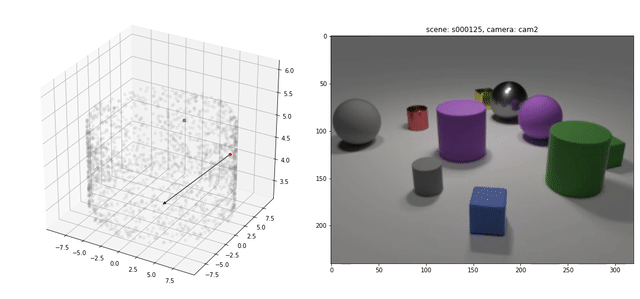}
        \caption{Two example views from one scene, with the camera position and vector plotted in 3D.}
        \label{fig:dataset_illustration_b}
    \end{subfigure}
    \caption{\ref{fig:dataset_illustration_a}: Schematic of the \clevrkiwi dataset. An image corresponding to the viewpoint camera is given as a query, but the corresponding question to answer is posed with respect to the question viewpoint. Given this image as well as the camera coordinates corresponding to the question viewpoint, the goal is to answer the question. \ref{fig:dataset_illustration_b}: Two example views from a single scene. In each example the corresponding camera position (vector) is shown. All other dots in the point cloud denote possible camera positions. To see a full gif animation of different camera views for this particular scene, please see \href{https://raw.githubusercontent.com/christopher-beckham/clevr-mrt-dataset-gen/kiwi_v3/out.gif}{here}.}
    \label{fig:dataset_illustration}
\end{figure}

The original CLEVR dataset provided one image for each scene. \emph{CLEVR-MRT} contains 20 images generated for each scene holding a constant altitude and sampling over azimuthal angle. To ensure that the model would not have any clues as to how the view had been rotated, we replaced the asymmetrical "photo backdrop" canvas of the CLEVR dataset with a large plane and centered overhead lighting. To focus on questions with viewpoint dependent answers, we filtered the set of questions to only include those containing spatial relationships (e.g. `is X to the right of Y'). From the original 90 question templates, only 44 contained spatial relationships. In total, the training + validation split consists of 45,600 scenes, each containing roughly 10 questions for a total of 455,549 questions. 5\% of these scenes were set aside for validation. For the test set, 10,000 scenes were generated with roughly 5 questions each, for a total of 49,670 questions. A schematic of the problem is illustrated in Figure \ref{fig:dataset_illustration}, and in Figure \ref{fig:overview}(b) we show a concrete example of one of these scenes.

\section{Methods}

We begin here by describing simple and strong baseline methods as well as upper bound estimates used to evaluate the performance of different techniques on this dataset. We then present our new approach to learning 3D features and two different ways to address this task.

\subsection{FILM baselines} \label{sec:film_baseline}

\begin{figure}[h]
    \centering
    \includegraphics[width=0.95\textwidth]{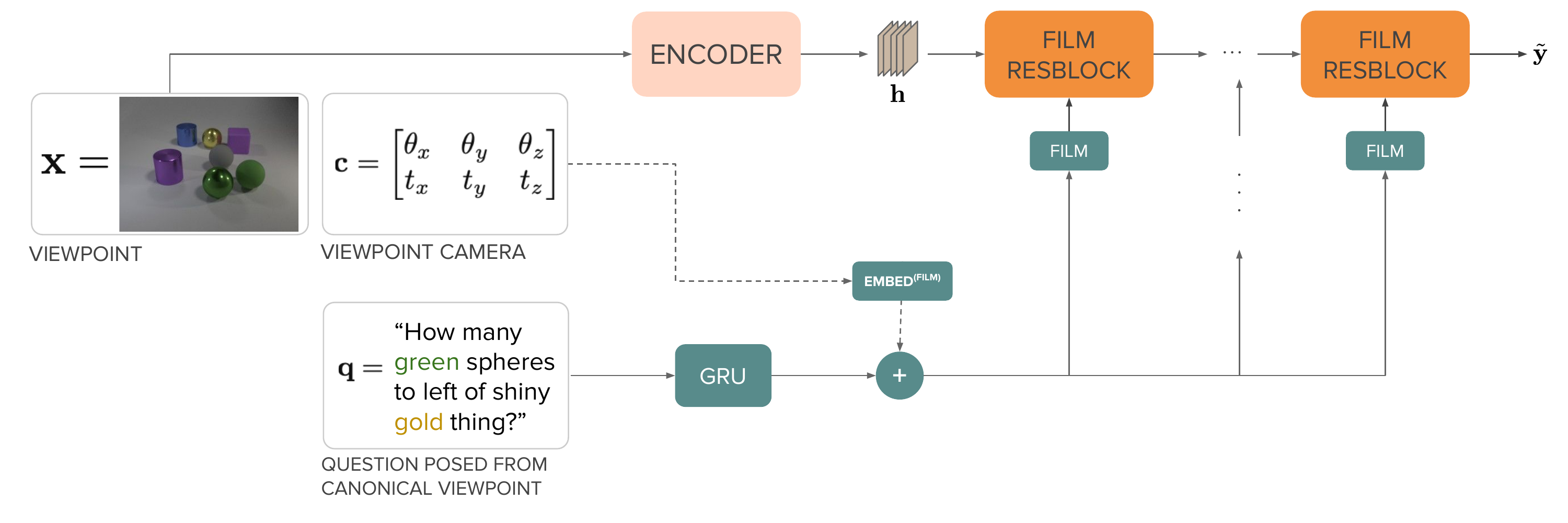}
    \caption{The pipeline of our FILM baselines \citep{film}. An input view (which is randomly sampled from a scene) is fed through a pre-trained encoder (a ResNet-101 pretrained on ImageNet) to produce a high-dimensional stack of feature maps $\mathbf{h}$ of dimension (1024, 14, 14). The question embedding is fed through a GRU which outputs an embedding vector of the sentence.  The viewpoint camera is also run through its own embedding module before being concatenated to the question embedding, and its arrows are presented as dashes to show that it is optional, depending on the precise baseline that is run. The resulting feature maps from the encoder are fed through FILM-modulated residual blocks using the final question/camera embedding vector. (Please see the supplementary materials section for more details.)}
    \label{fig:film_baseline}
\end{figure}

The architecture we use is based on FILM \citep{film}, in which a pre-trained ResNet-101 classifier on ImageNet extracts features from the input images which are then fed to a succession of FILM-modulated residual blocks using the hidden state output from the GRU. As a sanity check -- to ensure our models are adequately parameterised -- the simplest baseline to run is one where each scene in the dataset contains only one view: the \emph{canonical view}. In this setting, we would expect the highest validation performance since the canonical viewpoint is precisely the viewpoint that all questions are posed with respect to. The second and third baselines to run are ones where we use the \emph{full dataset}, with and without conditioning on the viewpoint camera via FILM, respectively. This is illustrated in Figure \ref{fig:film_baseline}, where we can see the viewpoint camera also being embedded before being concatenated to the question embedding and passed through the subsequent FILM blocks. If we let $S$ denote a scene consisting of all of its camera views (images) $\mathbf{X}$, the camera $\mathbf{c}$, the question $\mathbf{q}$, and its corresponding answer $\mathbf{y}$, we can describe the pipeline shown in Figure \ref{fig:film_baseline} as the following, with $\phi$ denoting the learnable parameters:

\noindent
\begin{align} \label{eq:film_baseline}
S = (\mathbf{X}, \mathbf{q}, \mathbf{c}, \mathbf{y}) & \sim \mathcal{D} \text{ (sample a scene)} \nonumber \\
\mathbf{x} & \sim \mathbf{X} \nonumber \text{  (sample a random view)}\\
\mathbf{h} &:= \text{encode}(\mathbf{x}) \nonumber \\
\mathbf{e}_{\text{cam}} &:= \text{embed}_{\phi}^{\text{(film)}}(\mathbf{c}) \nonumber \\
\mathbf{e}_{\text{gru}} &:= \text{GRU}_{\phi}(\mathbf{q}) \nonumber \\
\tilde{\mathbf{y}} &:= \text{FILM}_{\phi}(\mathbf{h}, [\mathbf{e}_{\text{gru}}, \mathbf{e}_{\text{cam}}]) \nonumber \\
\ell &:= \ell_{\text{cls}}(\mathbf{y}, \tilde{\mathbf{y}}),
\end{align}
where the encoder (pre-trained ResNet) is frozen and therefore has no learnable parameters. Here, $\ell_{\text{cls}}$ is the multinomial classification loss over the predicted answer token.

\subsection{Learning 3D Feature Representations from Single View Images}  \label{sec:film_in_3d}

So far we have been operating in 2D, based on the pre-trained ResNet-101 ImageNet encoder which outputs a high-dimensional stack of feature maps (a 3D tensor). To work in 3D, we would either need to somehow transform the existing encoding into a 4D tensor (a stack of 3D feature \emph{cubes}) or use a completely different encoder altogether which can output a 3D volume directly. Assuming we already had such a volume, we can manipulate the volume in 3D space directly by having it undergo any rigid transformation that is necessary for the question to be answered. In Section \ref{sec:proj_2d_to_3d} we illustrate a simple technique which simply takes the existing ImageNet encoder's features and runs it through a learnable `post-processing' block to yield a 3D volume, and in Section \ref{sec:contrastive} we propose a self-supervised contrastive approach to learn such an encoder from scratch without the use of camera extrinsics.

\subsubsection{Projecting 2D Features into 3D Features} \label{sec:proj_2d_to_3d}

\begin{figure}[h]
    \centering
    \includegraphics[width=0.9\textwidth]{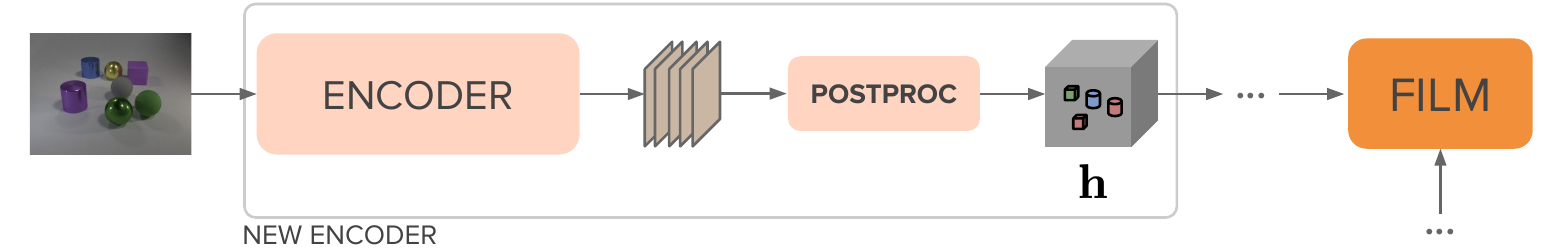}
    \caption{The pre-trained ResNet encoder outputs a stack of feature maps $\mathbf{h}$ of dimensions (1024, 14, 14), as in Fig. \ref{fig:film_baseline}, but now a post-processing module $\text{postproc}_{\psi}(\mathbf{h})$, e.g. a set of 2D convolutions, processes the feature stack and reshapes it into a 4D tensor $\mathbf{h}'$ of dimensions (64, 16, 14, 14) , i.e., a stack of feature \emph{cubes}. This entire block (inside the grey border) is the new `encoder', with the forward pass remaining the same as described in Equation \ref{eq:film_baseline}.}
    \label{fig:lifting}
\end{figure}

To exploit the power of pre-trained representations, we start with a pre-trained ResNet encoder and transform its stack of feature maps through an additional set of 2D convolution blocks using the `post-processor' shown in Figure \ref{fig:lifting}, right before reshaping the 3D tensor into 4D. In other words, we learn a module which maps from a stack of feature maps $\mathbf{h}$ to a stack of feature cubes, i.e. the encoding step in Equation \ref{eq:film_baseline} is replaced with $\mathbf{h} := \text{postproc}_{\phi}(\text{encode}(\mathbf{x}))$. Since the post-processor is a \emph{learnable} module through which the FILM part of the pipeline is able to backpropagate through, it can be seen as learning how to \emph{lift} said 2D representation into 3D. Through back-propagation it learns to perform well when manipulated with camera-conditioned FILM operations either as is or, more interestingly, when also subjected to rigid 3D transformations as we will see shortly in Section \ref{sec:camera_control_film}.

\subsubsection{3D Camera Controllable FILM} \label{sec:camera_control_film}

\begin{figure}[h!]
    \centering
    \includegraphics[width=0.95\textwidth]{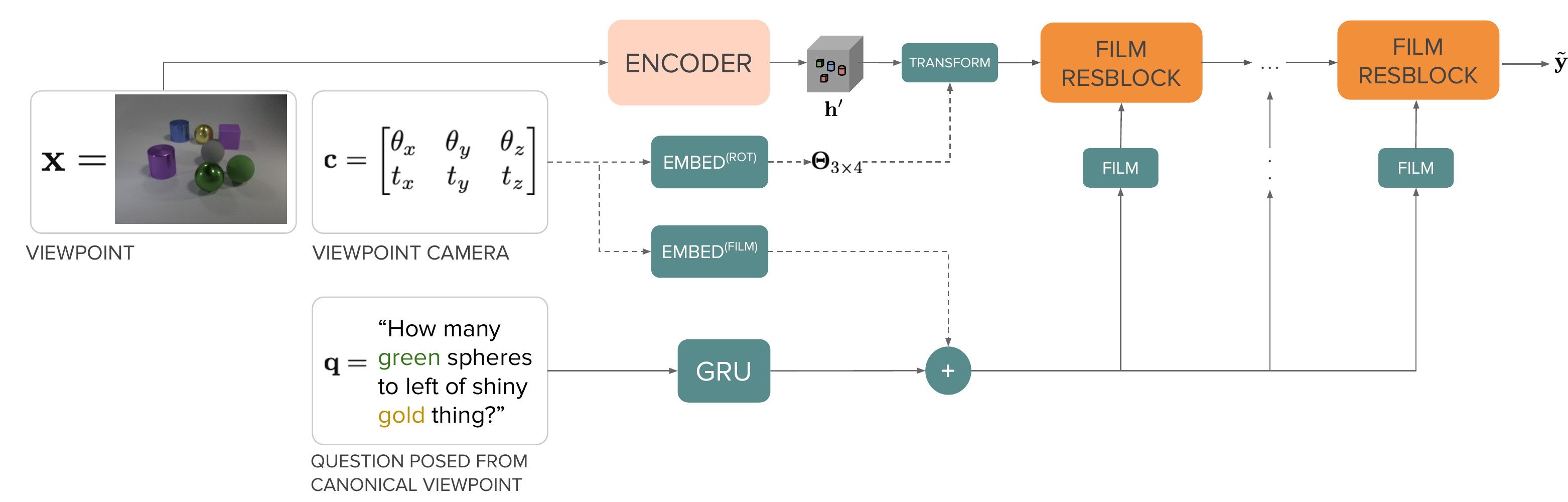}
    \caption{The 3D version of the FILM pipeline proposed in Section \ref{sec:film_in_3d}. The encoder can be either the 2D-to-3D formulation in Section \ref{sec:proj_2d_to_3d} (with the ResNet-101 inside it frozen but the postprocessor block learnable, i.e. Figure \ref{fig:lifting}) or the contrastive encoder in Section  \ref{sec:contrastive} (which directly maps to 3D volumes and is also frozen). A camera encoder \textbf{$\text{embed}_{\psi}^{\text{(rot)}}(\mathbf{c})$} is trained to map the camera coordinates of the scene to a transformation matrix which is used to transform the resulting 4D volume via an explicit rotation and translation, and/or it can be embedded and concatenated with the GRU embedding via \textbf{$\text{embed}_{\phi}^{\text{(film)}}$} (the dotted lines for both indicate that either/or are optional). This volume is then fed to FILM-modulated ResBlocks, which utilise 3D convolutions since $\mathbf{h}$ is now a volume instead of feature maps.}
    \label{fig:film_3d}
\end{figure}

In lieu of conditioning the camera with FILM (as seen in Figure \ref{fig:film_baseline} with $\text{embed}_{\phi}^{\text{(film)}}$), we can also condition on it to output translation and rotation parameters $(\tilde{\bm{\theta}}_{x}, \tilde{\bm{\theta}}_{y}, \tilde{\bm{\theta}}_{z}, \tilde{\mathbf{t}}_{x}, \tilde{\mathbf{t}}_{y}, \tilde{\mathbf{t}}_{z})$ which are then used to construct a 3D rotation and translation matrix $\Theta$ that $\mathbf{h}$ is subjected to (which is now a volume). Therefore, we can write out the 3D FILM pipeline as:
\begin{align} \label{eq:film_3d}
\mathbf{h} & := \text{postproc}_{\phi}(\text{encoder}(\mathbf{x})) \nonumber \\
(\tilde{\bm{\theta}}_{x}, \tilde{\bm{\theta}}_{y}, \tilde{\bm{\theta}}_{z}, \tilde{\mathbf{t}}_{x}, \tilde{\mathbf{t}}_{y}, \tilde{\mathbf{t}}_{z}) & := \text{embed}_{\phi}^{\text{(rot)}}(\mathbf{c}) \nonumber \\
\mathbf{h}_{\text{rot}} & := \text{transform}(\mathbf{h}; P(\tilde{\bm{\theta}}_{x}, \tilde{\bm{\theta}}_{y}, \tilde{\bm{\theta}}_{z}, \tilde{\mathbf{t}}_{x}, \tilde{\mathbf{t}}_{y}, \tilde{\mathbf{t}}_{z})) \nonumber \\
\tilde{\mathbf{y}} & := \text{FILM}_{\phi}(\mathbf{h}_{\text{rot}}, [\text{GRU}_{\phi}(\mathbf{q})]) \nonumber \\
\ell_{\text{cls}} & := \ell(\mathbf{y}, \tilde{\mathbf{y}}), 
\end{align}
where $P(\cdot)$ is a function that produces a rigid transform matrix from its arguments, which are Euler angles. This is illustrated in Figure \ref{fig:film_3d}.

Note that there are now two ways in which the viewpoint camera can modulate the VQA pipeline: either through FILM via $\text{embed}_{\phi}^{\text{(film)}}$ or by directly parameterising a rigid transformation with $\text{embed}_{\phi}^{\text{(rot)}}$. While both mechanisms are shown in Figure \ref{fig:film_3d}, for brevity's sake we have only shown in the latter in Equation \ref{eq:film_3d}. Also note that we cannot directly use the raw camera parameters $\mathbf{c} = (\bm{\theta}_{x}, \bm{\theta}_{y}, \bm{\theta}_{z}, \mathbf{t}_{x}, \mathbf{t}_{y}, \mathbf{t}_{z})$ to construct the rigid transform $\Theta$ because these are relative to world coordinates, not the canonical camera (whose coordinates are unknown). 

In the following next section (Section \ref{sec:contrastive}), we will show that we can replace the pre-trained ImageNet encoder and the learned postprocessor (Figure \ref{fig:lifting}) with a contrastive encoder trained from scratch.

\subsubsection{Learning 3D Contrastive Encoders}\label{sec:contrastive}

\begin{figure}[h!]
    \centering
    \includegraphics[width=0.9\linewidth]{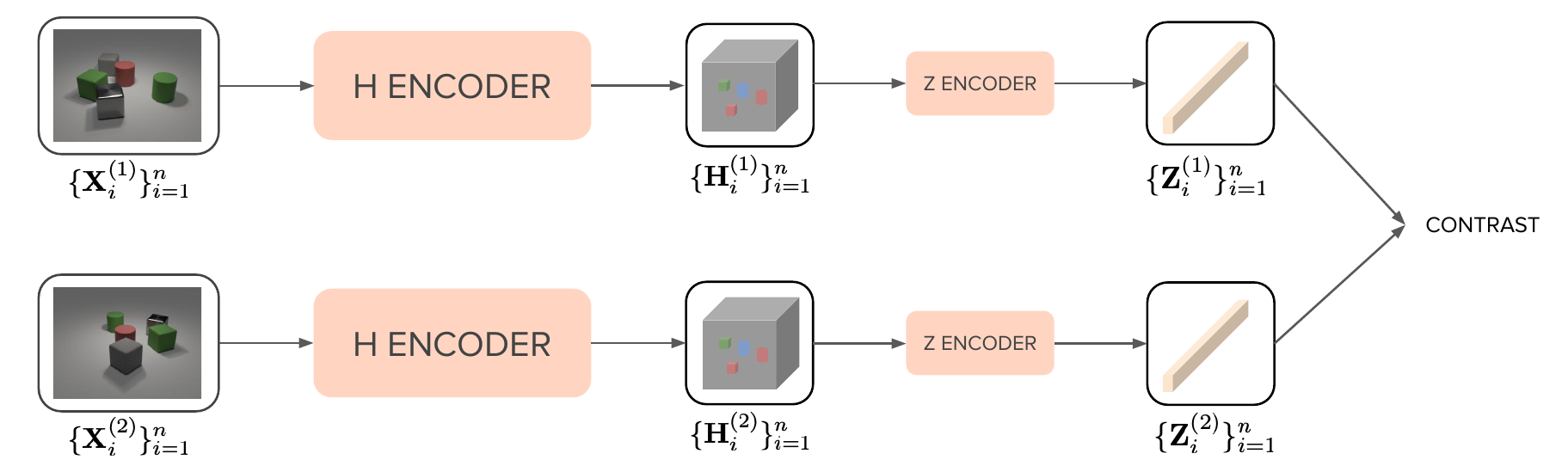}
    \caption{The contrastive-based encoder, inspired from \citep{contrastive_hinton}. We sample two sets of minibatches $\mathbf{X}^{(1)}$ and $\mathbf{X}^{(2)}$, where the $i$'th instance in each set comprises a positive pair (different views of the same scene). The $\mathbf{H}$ encoder generates a 3D volume for each view, and an additional $\mathbf{Z}$ encoder convolves this down to a summarisation vector over which the contrastive loss is applied.}
    \label{fig:contrastive}
\end{figure}

In Section \ref{sec:proj_2d_to_3d} the encoder proposed was an adaptation of a pre-trained ImageNet classifier backbone to output latent volumes. Here we propose the training of an encoder from scratch in an self-supervised manner, via the use of contrastive learning as demonstrated in \cite{contrastive_hinton}. Conceptually, we would like to learn a metric space where the distance between two views from the \emph{same} scene are minimised, and views from two \emph{different} scenes are maximised. In practice, the architecture we use is one which directly maps images to latent volumes via a sequence of 2D convolutions followed by 3D convolutions (the $\mathbf{h}$ encoder), followed by the $\mathbf{z}$ encoder which reduces those volumes down to latent codes which is what the contrastive loss operates on. This is shown in Figure \ref{fig:contrastive}. The key idea to note here is that the training of this encoder does not require any information about the cameras in the scene, nor labels describing objects in the scene (unlike with the ImageNet classifier that was repurposed as an encoder). While we obviously do use camera information for the VQA task itself (for instance the camera embedding modules in Figures \ref{fig:film_baseline} and \ref{fig:film_3d}), we stress that being able to pre-train an encoder as a separate step to the VQA task under limited label supervision is potentially very beneficial in real-world applications where obtaining labels for the VQA task is costly.

Let us denote $\mathbf{X}^{(1)}$ and $\mathbf{X}^{(2)}$ to be minibatches of images (views), with subscripts for individual examples in the minibatch (e.g. $\mathbf{X}^{(1)}_{j}$). We will assume that $(\mathbf{X}^{(1)}_{i}, \mathbf{X}^{(2)}_{j})$ correspond to the same scene if $i = j$, otherwise they are different. The InfoNCE loss \citep{oord2018representation} is defined as $\frac{1}{n}\sum_{i=1}^{n} \ell_{\text{NCE}}^{(i)}$:
\begin{align} \label{eq:contrastive}
\centering
\ell_{\text{NCE}}^{(i)} := - \log \frac{\text{exp}(\text{sim}(\mathbf{Z}^{(1)}_{i}, \mathbf{Z}^{(2)}_{i})/ \tau)}{\sum_{k=1}^{n} \text{exp}(\text{sim}(\mathbf{Z}^{(1)}_{i}, \mathbf{Z}^{(2)}_{k}) / \tau)},
\end{align}
where $\mathbf{H} = \enc_{\mathbf{h}}(T(\mathbf{X}))$, $\mathbf{Z} = \enc_{\mathbf{z}}(\mathbf{H})$, and $T(\cdot)$ is some stochastic data augmentation operator (e.g. random crops, flips, colour perturbations) which operates on each example independently in the batch. This loss also contains a temperature term $\tau$, which is a hyperparameter to optimise (in practice, we found $\tau = 0.1$ to produce the lowest softmax loss). Since a large number of negative examples is needed to learn good features, we train this encoder on an 8-GPU setup with a combined batch size of 2048. 

\begin{figure}[h]
    \centering
    \begin{subfigure}[b]{0.46\textwidth}
         \centering
         \includegraphics{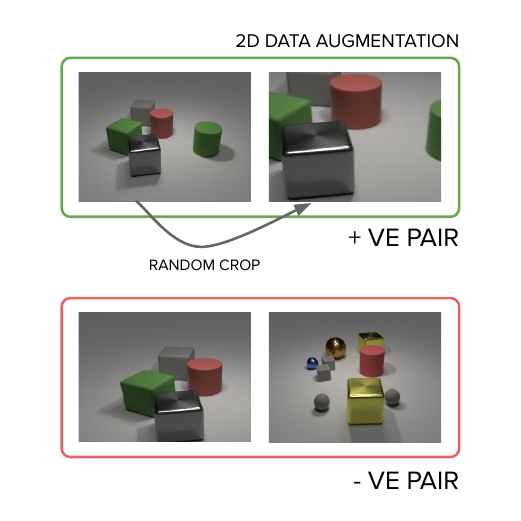}
         \caption{2D data augmentation}
         \label{fig:ctr_aug_2d}
    \end{subfigure}
    \begin{subfigure}[b]{0.46\textwidth}
        \centering
        \includegraphics{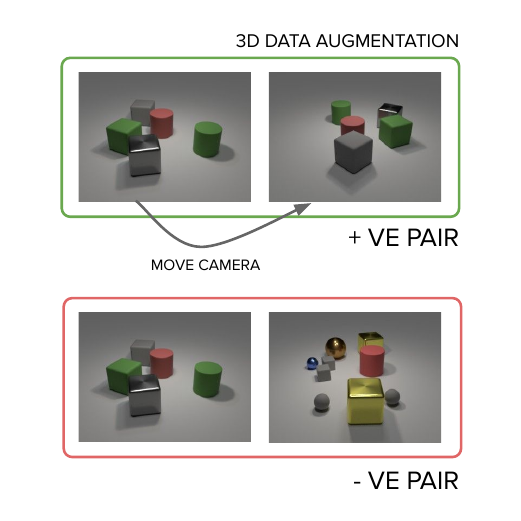}
        \caption{3D data augmentation}
        \label{fig:ctr_aug_3d}
    \end{subfigure}
    \par\bigskip
    \par\bigskip
    \begin{subfigure}[b]{0.55\textwidth}
        \centering
        \includegraphics{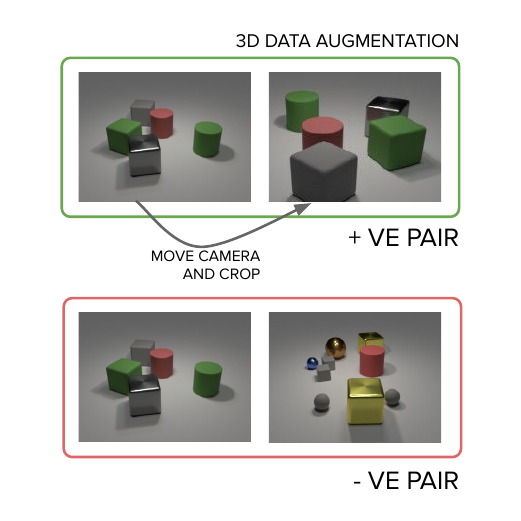}
        \caption{2D+3D data augmentation}
        \label{fig:ctr_aug_2d3d}
    \end{subfigure}
    \caption{\ref{fig:ctr_aug_2d}: 2D data augmentation. If the top left image of the positive pair is $\mathbf{X}^{(1)}$, then a positive pair is defined by $(T(\mathbf{X}^{(1)}), T(\mathbf{X}^{(1)}))$. In this illustration, $T$ is a stochastic function that produces random crops of an image with some pre-determined probability. \ref{fig:ctr_aug_3d}: 3D data augmentation. We leverage the fact that we have multiple views per scene which can be used to comprise a positive pair, and this can be denoted as simply $(\mathbf{X}^{(1)}, \mathbf{X}^{(2)})$, where $\textbf{X}^{(2)}$ comes from the same scene as $\textbf{X}^{(1)}$. \ref{fig:ctr_aug_2d3d}: Furthermore, both 2D and 3D data augmentation can be employed simultaneously, and this is required in order to pre-train an encoder that works well for VQA, as shown in Table \ref{tb:results_ctr}.}
    \label{fig:ctr_aug}
\end{figure}

Note that for the datasets considered in \cite{contrastive_hinton} (for instance ImageNet and CIFAR-10), positive pairs are generally stochastic 2D data augmentations of the \emph{same} image, i.e. ($T(\mathbf{X}), T(\mathbf{X}))$ comprises a positive pair and $T(\cdot)$ is stochastic. In our case, since our dataset consists of scenes which in turn consist of many views, we can easily construct a positive pair by sampling two different views from the same scene. This can be thought of as a form of `3D' data augmentation where instead of relatively primitive operations like crops and colour perturbations we are moving a camera around a scene and re-rendering it. We can also choose to perform both 2D and 3D data augmentation to ensure maximise diversity of positive pairs. To make things clear later, we define the positive pair $(T(\mathbf{X}^{(1)}), T(\mathbf{X}^{(2)}))$ as `2D only' data augmentation if $\mathbf{X}^{(1)} = \mathbf{X}^{(2)}$. If $\mathbf{X}^{(1)} \neq \mathbf{X}^{(2)}$ then this is `2D + 3D' data augmentation, and if $T$ is simply the identity function then this corresponds to `3D only' data augmentation, which is the pair $(\mathbf{X}^{(1)}, \mathbf{X}^{(2)})$. This is shown in Figure \ref{fig:ctr_aug}.

\section{Results and Analysis}\label{sec:results}

The pre-trained ImageNet backbone we use is the one that is pre-packaged with the PyTorch \texttt{torchvision} module, which is a ResNet-101 \cite{resnets}. For the 3D contrastive encoder (Figure \ref{fig:contrastive}), the backbone is simply a sequence of strided 2D Conv-BN-ReLU blocks, the result of which is subsequently reshaped into a 3D volume (a 4D tensor) and post-processed with 3D convolution blocks which output the final 3D latent volume. For the pre-training of this, the contrastive loss operates on the flat vector representation of the latent volume, which is simply computed with average pooling over the spatial axes. For the FILM pipeline, we use a sequence of ResNet blocks with ReLU nonlinearities and batch normalisation, as well as CoordConv feature maps \cite{coordconv}. Each experiment was trained for a maximum of 60 epochs with the ADAM optimiser \cite{adam}, with a default learning rate of $3 \times 10^{-4}$ and first and second moment coefficients $\{\beta_{1}, \beta_{2}\} = \{0.9, 0.999\}$.

For each experiment, we perform a sweep over many hyperparameters of the FILM architecture (detailed in Table \ref{tb:hps}) to find the experiment which performs the best, according to validation set accuracy. We then select the best-performing experiment and run repeats of it with varying seeds (3-6, depending on the variance exhibited by those experiments), for a total of $N=6$ runs. The validation set accuracy reported is the mean over these $N$ runs, and similarly for the test set. It is worth noting that for our postprocessor experiments (Section \ref{sec:proj_2d_to_3d}), some runs appeared to hit undesirable local minima, exhibiting much lower validation accuracies. We conjecture is due to a `domain mismatch' between our dataset and ImageNet, and this conjecture appears to be supported by the fact that these outliers do not exist when we use our pre-trained contrastive encoder (Section \ref{sec:contrastive}). To deal with these outliers, we instead compute the mean/stdev over only the top three runs out of the $N = 6$ we originally trained.

\begin{table}[h]
    \centering
    \caption{Table of results for experiments run using a pre-trained ResNet-101 encoder. The \emph{Upper bound} model is a baseline model where \emph{only} canonical views are given as input, and is expected to have the highest performance since it does not have to answer questions from another random viewpoint as input. For the columns shown: \textbf{3D?} refers to whether we are using 2D latents or 3D latents (the difference between Figs \ref{fig:film_baseline} and \ref{fig:film_3d}); \textbf{camera (embed)} refers to embedding the camera coordinates and concatenating it with the question embedding; \textbf{camera (rotation)} refers to using the camera to map to a rigid transform of the volume (shown in Figure \ref{fig:film_3d}). The result denoted with $\bm{\dagger}$ (in small text) indicates the same experiment but with the postprocessor frozen after random weight initialisation. See Appendix Section \ref{supp:mac} for more on MAC baseline. Accuracies shown are percentages. For all \emph{3D FILM, projection} results, the mean/stdev is computed over the top 3 performing models (out of 6).}
    \label{tb:results}
    \begin{tabular}{ >{\raggedright\arraybackslash}p{4.0cm}  >{\centering\arraybackslash}m{0.5cm} >{\centering\arraybackslash}m{1.5cm} >{\centering\arraybackslash}m{1.5cm} >{\raggedleft\arraybackslash}m{2.3cm} >{\raggedleft\arraybackslash}m{2.3cm} } 
    \toprule
    Method & 3D? & camera (embed) & camera (rotation) & valid acc. (\%)  & test acc. (\%)   \\ 
    \midrule
    Majority class & -- & -- & -- & 24.72 $\pm$ 0.00 & 24.75 $\pm$ 0.00 \\
    GRU-only & -- & \xmark & \xmark & 49.38 $\pm$ 0.40 & -- \\
    Upper bound (\emph{canon. views only}) & \xmark & \xmark & \xmark & 94.19 $\pm$ 0.39 & 94.24 $\pm$ 0.40 \\
    \midrule
    MAC \citep{mac} & \xmark & \cmark & \xmark & 70.96 $\pm$ 0.97 & -- \\
    \midrule
    \multirow{2}{*}{2D FILM (Sec \ref{sec:film_baseline}, Fig \ref{fig:film_baseline}) } & \xmark & \xmark & \xmark & 70.63 $\pm$ 0.19 & 69.60 $\pm$ 0.09 \\
        & \xmark & \cmark & \xmark & 83.95 $\pm$ 1.21 & 83.68 $\pm$ 1.21 \\
    \midrule
    \multirow{4}{*}{\parbox{4.1cm}{3D FILM, projection (Sec \ref{sec:proj_2d_to_3d}, Fig \ref{fig:film_3d})} } & \cmark & \xmark & \xmark & 68.19 $\pm$ 1.87 & -- \\
        & \cmark & \cmark & \xmark & 88.82 $\pm$ 3.04 & 86.36 $\pm$ 3.46 \\
        & \cmark & \xmark & \cmark & \parbox{2.1cm}{\raggedleft{\textbf{92.80 $\pm$ 0.30}} \\ \scriptsize{ ($\dagger$ 68.98 $\pm$ 1.43) }} & \textbf{ 90.86 $\pm$ 0.87 } \\
        & \cmark & \cmark & \cmark & 89.83 $\pm$ 1.36 & 89.68 $\pm$ 1.34 \\
    \bottomrule
    \end{tabular}
\end{table}

\begin{table}[t]
    \centering
    \caption{Table of results for experiments run with the contrastive encoder described in Section \ref{sec:contrastive}. For the columns shown: \textbf{Data aug} refers to what data augmentation scheme was used (see Figure \ref{fig:ctr_aug}): \emph{2D} = +ve pair is  $T(\mathbf{X}^{(1)})$ and $T(\mathbf{X}^{(1)})$, \emph{3D} = +ve pair is $\mathbf{X}^{(1)}$ and $\mathbf{X}^{(2)}$, \emph{2D+3D} = +ve pair is $T(\mathbf{X}^{(1)})$ and $T(\mathbf{X}^{(2)}$); \textbf{NCE accuracy} is how well the contrastive encoder is able to distinguish between pairs of views that belong to the same/different scene; the remaining columns denote the FILM task, as described in Table \ref{tb:results}.}
    \label{tb:results_ctr}
    \begin{tabular}{ >{\raggedright\arraybackslash}p{1.5cm}  >{\centering\arraybackslash}m{0.3cm} >{\centering\arraybackslash}m{2.0cm} >{\centering\arraybackslash}m{1.4cm} >{\centering\arraybackslash}m{1.4cm} >{\raggedleft\arraybackslash}m{2.2cm} >{\raggedleft\arraybackslash}m{2.2cm} } 
    \toprule
    \multicolumn{3}{c}{Contrastive pre-training stage} & \multicolumn{4}{c}{FILM stage} \\
    \cmidrule(lr){1-3} \cmidrule(lr){4-7} 
    Data aug & $\tau$ & NCE accuracy (valid) & camera (embed) & camera (rotation) & valid acc. (\%) & test acc. (\%) \\
    \toprule
    \multirow{2}{*}{2D} & \multirow{2}{*}{0.1} &  \multirow{2}{*}{9.13} &  \cmark &  \xmark & 59.78 $\pm$ 0.23 & 59.14 $\pm$ 0.43 \\
    &  &  &  \xmark &  \cmark & 59.29 $\pm$ 0.44 & 58.57 $\pm$ 0.53 \\
    \midrule
    \multirow{2}{*}{3D} & \multirow{2}{*}{0.1} & \multirow{2}{*}{99.72} &  \cmark &  \xmark & 57.42 $\pm$ 0.26 & 56.74 $\pm$ 0.31 \\
    &  &  &  \xmark &  \cmark & 57.63 $\pm$ 0.21 & 57.10 $\pm$ 0.33 \\
    \midrule
    \multirow{2}{*}{2D + 3D} & \multirow{2}{*}{0.1} & \multirow{2}{*}{98.14} & \cmark & \xmark & 65.15 $\pm$ 4.63 & 63.70 $\pm$ 3.74 \\
    &  &  & \xmark & \cmark & \textbf{87.49 $\pm$ 0.78} & \textbf{86.01 $\pm$ 0.69} \\
    \bottomrule
    \end{tabular}
\end{table}

Our results for the FILM baselines (Section \ref{sec:film_baseline}) and using 2D-to-3D projections (Section \ref{sec:proj_2d_to_3d}) are shown in Table \ref{tb:results}. What we find surprising is that the 2D baseline without camera conditioning (top-most row of \emph{2D FILM}) is able to achieve a decent accuracy of roughly 70\%. On closer inspection the misclassifications do not appear to be related to how far away the viewpoint camera is from the canonical, with misclassified points being distributed more or less evenly around the scene. Given that the question-only baseline (`GRU-only') is able to achieve an accuracy significantly greater than that of the majority class baseline ($\approx 25\%$ versus $\approx 50\%$), it seems like it is likely exploiting statistical regularities between the actual question itself and the scene. If we add camera conditioning via FILM (bottom-most row of \emph{2D FILM}) then we achieve a much greater test accuracy of $83.68 \pm 1.21$. Furthermore, as shown in the \emph{3D FILM} part of Table \ref{tb:results}, our results demonstrate the efficacy of using rigid transforms with 3D volumes, achieving the highest accuracy of $90.86 \pm 0.87$ on the test set (highlighted in bold). If we take the same experiment and freeze the postprocessor (denoted by the small $\dagger$ symbol), then we achieve a much lower accuracy of 69 \%. This is to be expected, considering that any camera information that is forward-propagated will contribute gradients back to the postprocessing parameters in the backward propagation, effectively giving the postprocessor supervision in the form of camera extrinsics.  Finally, the last row of the \emph{3D FILM} shows that if one uses the camera for both rigid transforms \emph{and} embedding, the mean test accuracy is roughly the same as the rigid-transform-only variant (90.86 $\pm$ 0.87 vs 89.68 $\pm$ 1.34).  This appears to suggest that simply performing rigid rotations of the volume is sufficient by itself for good performance.

\begin{table}[h]
    \centering
    \caption{Select experiments from Table \ref{tb:results} but trained on CLEVR-MRT-v2 (Figure \ref{fig:clevr_mrt_example}) where small objects exist and camera elevation is allowed to vary. For all experiments shown in this table, the mean/stdev is computed over the top three runs (out of six in total).}
    \label{tb:results_mrt}
    \begin{tabular}{ >{\raggedright\arraybackslash}p{4.3cm}  >{\centering\arraybackslash}m{0.5cm} >{\centering\arraybackslash}m{1.4cm} >{\centering\arraybackslash}m{1.4cm} >{\raggedleft\arraybackslash}m{2.2cm} >{\raggedleft\arraybackslash}m{2.2cm} } 
    \toprule
    Method & 3D? & camera (embed) & camera (rotation) & valid acc. (\%)  & test acc. (\%)   \\ 
    \midrule
    Upper bound (\emph{canon. views only}) & \xmark & \xmark & \xmark & 90.00 $\pm$ 0.23 & 89.37 $\pm$ 0.19 \\
    \midrule
    \multirow{2}{*}{2D FILM (Sec \ref{sec:film_baseline}, Fig \ref{fig:film_baseline}) } & \xmark & \xmark & \xmark & 67.26 $\pm$ 0.78 & -- \\
    & \xmark & \cmark & \xmark & 79.69 $\pm$ 2.05 & 79.14 $\pm$ 2.35 \\
    \midrule
    \multirow{3}{*}{\parbox{4.1cm}{3D FILM, projection (Sec \ref{sec:proj_2d_to_3d}, Fig \ref{fig:film_3d})} } &  \cmark & \cmark & \xmark & 65.49 $\pm$ 1.46 & 65.10 $\pm$ 1.67 \\
        & \cmark & \xmark & \cmark & 86.92 $\pm$ 2.00 & 86.89 $\pm$ 2.04 \\
        & \cmark & \cmark & \cmark & \textbf{89.98 $\pm$ 0.59} & \textbf{89.91 $\pm$ 0.73} \\
    \bottomrule
    \end{tabular}
\end{table}

In Table \ref{tb:results_ctr} we present 3D FILM results but using the contrastive pre-trained encoder described in Section \ref{sec:contrastive}. Specifically, we perform an ablation on the type of data augmentation used during the contrastive pre-training stage (described at the end of Section \ref{sec:contrastive}) and find that 3D data augmentation is essential for the encoder to distinguish whether a pair of views come from the same scene or not, as shown in the `NCE accuracy' column (9.13\% for 2D versus 99.72 \% for 3D). However, both 2D and 3D data augmentation is necessary in order for the FILM task to yield the best results, as seen in the last row. This is consistent with the observation that very strong data augmentation is required to ensure that contrastive techniques do not learn trivial features that perform poorly on downstream tasks. Similar to Table \ref{tb:results}, utilising the viewpoint camera for rigid transforms produces the best results, with 86.01 $\pm$ 0.69 \% test accuracy. While the best result of Table \ref{tb:results} is slightly higher, we re-iterate that some of those runs hit undesirable local minima, which we did not experience with this contrastive formulation. Furthermore, as noted in \cite{contrastive_hinton}, contrastive encoders have to be significantly overparameterised compared to their supervised counterparts in order to achieve roughly the same classification error, so further architecture tuning may be required.

\begin{figure}[h]
     \centering
     \includegraphics[width=0.19\textwidth]{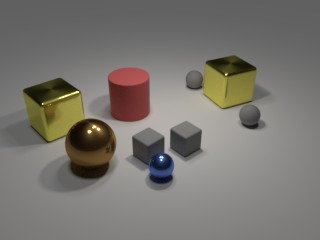}
     \includegraphics[width=0.19\textwidth]{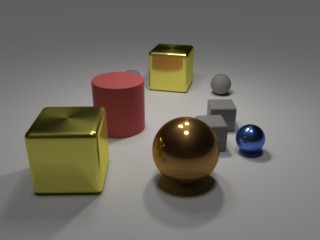}
     \includegraphics[width=0.19\textwidth]{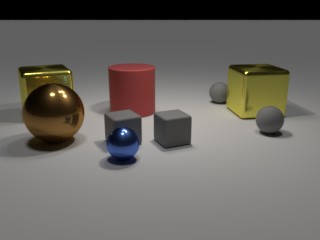}
     \includegraphics[width=0.19\textwidth]{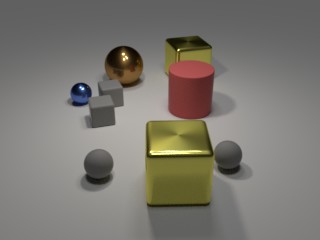}
     \includegraphics[width=0.19\textwidth]{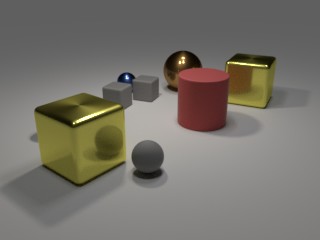}
     \par\bigskip
     \includegraphics[width=0.19\textwidth]{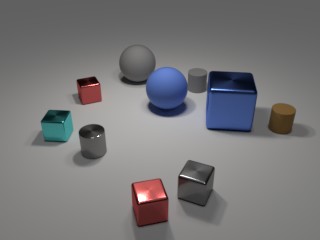}
     \includegraphics[width=0.19\textwidth]{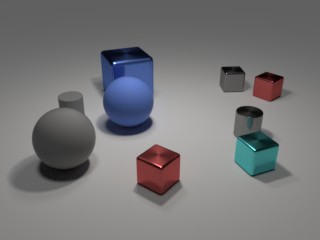}
     \includegraphics[width=0.19\textwidth]{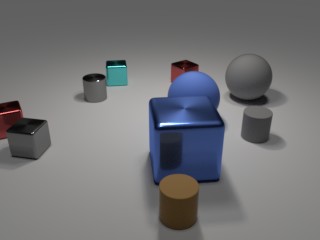}
     \includegraphics[width=0.19\textwidth]{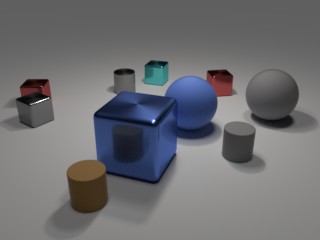}
     \includegraphics[width=0.19\textwidth]{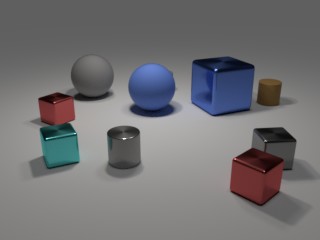}
     \par\bigskip
     \includegraphics[width=0.19\textwidth]{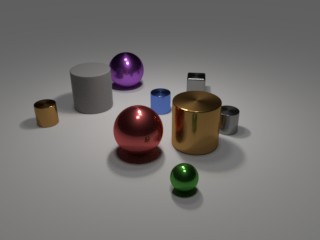}
     \includegraphics[width=0.19\textwidth]{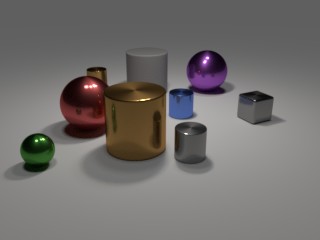}
     \includegraphics[width=0.19\textwidth]{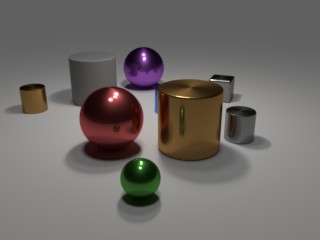}
     \includegraphics[width=0.19\textwidth]{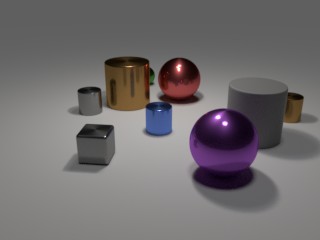}
     \includegraphics[width=0.19\textwidth]{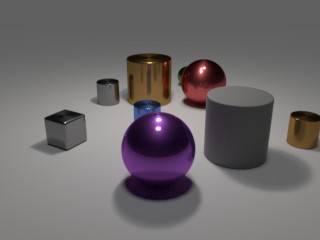}
     \caption{\clevrkiwi \ v2 dataset. For each scene (corresponding to a row), the left-most image is the canonical view. The rest are randomly sampled views, of varying azimuth and elevation, the latter of which is now allowed to vary, unlike the original CLEVR-MRT dataset. Because of this, occlusions here are more prevalent compared to the original dataset shown in Figure \ref{fig:overview_b}.}
     \label{fig:clevr_mrt_example}
\end{figure}

Our results show that our best performing formulation (either 2D-to-3D or contrastive) performs on average only 8\% less than the canonical baseline's 94\%, which can be seen as a rough upper bound on generalisation performance. While we obtained promising results, it may not leave a lot of room to improve on top of our methods, and we identified some ways in which the dataset could be made more difficult. One of them is removing the viewpoint camera coordinates and instead placing a sprite in the scene showing where the canonical camera is. This means that the model has an additional inference task it has to perform, which is inferring the 3D position of the canonical viewpoint from a 2D marker. Another idea is to allow some variation in the elevation of the viewpoint camera. While this can accentuate the effects of occlusion (if the camera is allowed to go lower than its default elevation), it also provides for a more grounded dataset since occlusions are commonplace in real-world datasets. We examine the latter here, generating a version of CLEVR-MRT where the camera elevation is allowed to vary, and with both small and large objects present in the scene (small objects were not present in the original CLEVR-MRT dataset). The default elevation in the original dataset was $30^{\circ}$, whereas now it is randomly sampled from $\text{Uniform}(20, 30)$. An example scene of this new dataset, CLEVR-MRT-v2, is shown in Figure \ref{fig:clevr_mrt_example}.

See Table \ref{tb:results_mrt} for these results. We also demonstrate here that 3D FILM performs the best, though compared to Table \ref{tb:results} it appears that camera conditioning via FILM is also required for a few extra percentage points (bottom-most row of \emph{3D FILM}). Like Table \ref{tb:results}, it appears that our best result is on par with the upper bound, indicating that more difficult versions of the dataset are required. As we mentioned earlier, one addition would be to allow the camera to be present in the scene as a model or a sprite, with the goal of inferring the coordinates of the camera to answer the question without the viewpoint camera being provided as it is now. This would also have the benefit of making the dataset more realistic with respect to the illustrative examples given in the introduction. We leave this to future work however.

\subsection{Limitations and Future Work}

\begin{figure}[h]
    \centering
    \begin{subfigure}[b]{0.35\textwidth}
         \centering
         \includegraphics[width=\textwidth]{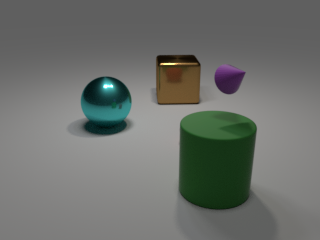}
         \caption{}
         \label{fig:clevr_mrt_v3_poc_example}
    \end{subfigure} \ \
    \begin{subfigure}[b]{0.35\textwidth}
        \centering
        \includegraphics[width=\textwidth]{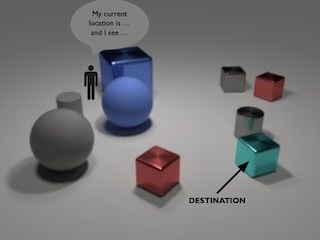}
        \caption{}
        \label{fig:clevr_mrt_poc_robotnav}
     \end{subfigure}
    \caption{\ref{fig:clevr_mrt_v3_poc_example}: Visualization of a more difficult formulation of \clevrkiwi \ where the camera is now visible as floating purple cone. Instead of being given its coordinates a-priori to condition on the rotation of the latent volume, its coordinates will also have to be inferred from the image. \ref{fig:clevr_mrt_poc_robotnav}: An example use-case for systems capable of performing CLEVR-MRT tasks. A system that is able to answer questions in natural language about what would be seen from another viewpoint, could be leveraged to provide navigation instructions using natural language to an agent seeking a destination or object. In this scenario the agent wishes to navigate to the shiny cyan cube. A system could provide instructions of the form: \textbf{\textit{`To find the shiny cyan cube go straight past the blue sphere on your left and the grey sphere on your right until you are in front of the red cube. You should see the cyan cube to your left as you approach the red cube.'}} Multiple rounds of interaction may be needed, but a key capability here is the ability of the system to understand what would be seen from a particular viewpoint.}
    \label{fig:clevr_mrt_future_examples}
\end{figure}

We have examined a \clevrkiwi \  setup where one is given an image as well as coordinates of the question viewpoint. Other tasks involving mental rotations might require the question viewpoint to be inferred by the neural network. For instance, if we wish to infer what another agent sees, we might wish to infer the position and orientation of their face or camera. One might reformulate our setup to include some sort of marker or object representing the desired camera position and orientation, and task the model with inferring that position and orientation. In Figure \ref{fig:clevr_mrt_v3_poc_example} we visualize this scenario where the desired camera position and viewpoint is illustrated in the scene as a purple cone. In this scenario instead of the VQA pipeline being given camera coordinates, they must inferrred from the appearance of the code, in addition to performing the mental rotation required to answer the question. While we leave such a task to future research we plan to release our formulation of CLEVR-MRT as a dataset in our code repository to encourage future work on more challenging mental rotation based tasks. 

Another interesting and related task is that of using natural language to guide robot navigation, and another version of CLEVR for this task has been proposed and examined in \cite{clevr_avs} within a reinforcement learning setup. In \cite{clevr_avs}, each scene contains a large occluding object in the center of the scene and the agent must try to navigate around it in order to answer the question. However, in the dataset of \cite{clevr_avs} it appears that the occluding object is always a large object in the center of the scene, which limits the variability of the scenes. Conversely, our dataset has significantly more occlusion variability when the camera's elevation is such that it is very close to the ground (see Figure \ref{fig:clevr_mrt_example}). Therefore, ideas from both these datasets could be combined to construct a more difficult CLEVR-like dataset specifically targeted towards navigation tasks.

Our work also has applications in indoor navigation for humans, using natural language to assist themn. Consider a user trying to navigate to a particular location in an unknown building while receiving instructions from a neural network acting as a navigation system. In such an indoor scenario, occlusion is extremely commonplace: potted plants, desks, chairs, walls, and doors. Smaller objects such as desks and chairs may not necessarily occupy a static position, and may change location. Unfamiliar room locations may be hard to find. As we discussed in Section \ref{sec:related}, for these kinds of tasks it is not strictly necessary to perform a pixel-level 3D reconstruction of the scene, but rather powerful neural representations that can be modulated through natural language. With an understanding of what a user would see at different locations, a system could better provide navigation assistance.

In Figure \ref{fig:clevr_mrt_poc_robotnav} we also provide a concrete example of how learning to perform the CLEVR-MRT tasks is related to the types of language guided navigation applications discussed above. By understanding what would be seen by an agent at a particular location and viewing position, a system could better formulate natural language instructions for how to reach a desired destination. Our current system is limited in that we have not implemented a mechanism to transform the answers to questions about alternative viewpoints into sequences of navigation instructions, but we feel that integrating CLEVR-MRT approaches into these navigation scenarios would be an interesting direction for future research.


In terms of limitations related to our contrastive learning results, we note that the particular algorithm we used, SimCLR \cite{contrastive_hinton}, requires a large batch size so that many negative examples can be contrasted against per minibatch. Such a large batch size can be prohibitively expensive in practice and require GPUs with large amounts of memory.  Other recent work that alleviates this includes MoCO \cite{moco} and SimSiam \cite{simsiam}, the latter of which does not require any negative examples whatsoever. 

Lastly, another limitation of our work concerns the interpretability of the learned latent feature volumes. Although we argue that certain tasks such as VQA do not necessarily require an explicit 3D reconstruction of the scene (see Section \ref{sec:related}), it may be beneficial from an interpretability or explainability point of view to consider an additional neural rendering step which takes the latent volume as input and re-renders the scene, possibly in another viewpoint. This can certainly be useful as a diagnostic tool to probe learned neural representations.

\section{Conclusion}

We address the problem of answering a question from a single image, posed in a reference frame that is different to the one of the viewer. We illustrate the difficulties here in a controlled setting, proposing a new CLEVR dataset and exploring a 3D FILM-based architecture that operates directly on latent feature volumes (using camera conditioning via FILM or via direct rigid transforms on the volume). We propose two techniques to train volumetric encoders: with 2D-to-3D projection of pre-trained ImageNet features, or using a self-supervised contrastive approach. In the latter case, we showed that the use of combined 2D+3D data augmentation was crucial to learning a volumetric encoder, as well as performing almost just as well as pre-trained ImageNet features. Because pre-training such a self-supervised encoder does not require supervision in the form of camera extrinsics or dataset-specific labels, it is more economically feasible for datasets where rich labels are not available. Through rigorous ablations, we demonstrated that performing 3D FILM was the most effective for \clevrkiwi, especially when the latent volume can be subjected to rigid transformations in order to answer the question. While the efficacy of our method has been demonstrated empirically, we identified several avenues in which \clevrkiwi \ can be made more realistic and challenging. Some examples of this include: an additional task of inferring the camera from the scene before performing the mental rotation; robot navigation tasks with a large variety of occluding objects; and language-guided indoor navigation. Lastly, while the use of an artificially-created dataset like CLEVR makes it easy to probe and test various properties of our algorithms, real-world VQA datasets involving mental rotations are required for a more practical validation of our proposed methods.

\subsection{Broader Impacts}

Endowing intelligent embodied systems with the ability to answer questions regarding properties of a 3D visual scene with respect to the perspective of another agent could make such systems safer. In the case of an autonomous vehicles, better control decisions could eventually be made. If such systems are adversarial in nature, negative outcomes could arise to the adversaries of such systems.

\section*{Acknowledgements}
The authors are grateful to the Natural Sciences and Engineering Research Council (NSERC) of Canada, and PROMPT Quebec for their support of this work. The first author also thanks the Mathematics of Information Technology and Complex Systems (MITACS) organization as well as the Institute for Data Valorization (IVADO) for their support. We also thank the Canadian Institute for Advanced Research (CIFAR) for their support under the Artificial Intelligence Research Chairs program. The primary author is grateful to David Vasquez and Catherine Martin for facilitating a research internship at ServiceNow Research.

\bibliography{biblio}

\begin{thebibliography}{47}
\providecommand{\natexlab}[1]{#1}
\providecommand{\url}[1]{\texttt{#1}}
\expandafter\ifx\csname urlstyle\endcsname\relax
  \providecommand{\doi}[1]{doi: #1}\else
  \providecommand{\doi}{doi: \begingroup \urlstyle{rm}\Url}\fi

\bibitem[Bachman et~al.(2019)Bachman, Hjelm, and Buchwalter]{amdim}
Philip Bachman, R~Devon Hjelm, and William Buchwalter.
\newblock Learning representations by maximizing mutual information across
  views.
\newblock In \emph{{Advances in Neural Information Processing Systems}}, pp.\
  15535--15545, 2019.

\bibitem[Bahdanau et~al.(2019)Bahdanau, de~Vries, O'Donnell, Murty, Beaudoin,
  Bengio, and Courville]{closure}
Dzmitry Bahdanau, Harm de~Vries, Timothy~J O'Donnell, Shikhar Murty, Philippe
  Beaudoin, Yoshua Bengio, and Aaron Courville.
\newblock {CLOSURE:} assessing systematic generalization of {CLEVR} models.
\newblock In \emph{{Visually Grounded Interaction and Language (ViGIL)
  workshop, NeurIPS 2019}}, 2019.

\bibitem[Behjati et~al.(2023)Behjati, Rodriguez, Fern{\'a}ndez, Hupont, Mehri,
  and Gonz{\`a}lez]{srres_behjati2023single}
Parichehr Behjati, Pau Rodriguez, Carles Fern{\'a}ndez, Isabelle Hupont, Armin
  Mehri, and Jordi Gonz{\`a}lez.
\newblock Single image super-resolution based on directional variance attention
  network.
\newblock \emph{{Pattern Recognition}}, 133:\penalty0 108997, 2023.

\bibitem[Chen et~al.(2020)Chen, Kornblith, Norouzi, and
  Hinton]{contrastive_hinton}
Ting Chen, Simon Kornblith, Mohammad Norouzi, and Geoffrey Hinton.
\newblock A simple framework for contrastive learning of visual
  representations.
\newblock In \emph{{International Conference on Machine Learning}}, pp.\
  1597--1607. PMLR, 2020.

\bibitem[Chen \& He(2021)Chen and He]{simsiam}
Xinlei Chen and Kaiming He.
\newblock Exploring simple siamese representation learning.
\newblock In \emph{Proceedings of the IEEE/CVF Conference on Computer Vision
  and Pattern Recognition}, pp.\  15750--15758, 2021.

\bibitem[Dou et~al.(2018)Dou, Wu, Shah, and
  Kakadiaris]{srface_dou2018monocular}
Pengfei Dou, Yuhang Wu, Shishir~K Shah, and Ioannis~A Kakadiaris.
\newblock Monocular {3D} facial shape reconstruction from a single {2D} image
  with coupled-dictionary learning and sparse coding.
\newblock \emph{{Pattern Recognition}}, 81:\penalty0 515--527, 2018.

\bibitem[Eslami et~al.(2018)Eslami, Rezende, Besse, Viola, Morcos, Garnelo,
  Ruderman, Rusu, Danihelka, Gregor, et~al.]{gqn}
SM~Ali Eslami, Danilo~Jimenez Rezende, Frederic Besse, Fabio Viola, Ari~S
  Morcos, Marta Garnelo, Avraham Ruderman, Andrei~A Rusu, Ivo Danihelka, Karol
  Gregor, et~al.
\newblock Neural scene representation and rendering.
\newblock \emph{Science}, 360\penalty0 (6394):\penalty0 1204--1210, 2018.

\bibitem[Fahim et~al.(2021)Fahim, Amin, and Zarif]{fahim2021single}
George Fahim, Khalid Amin, and Sameh Zarif.
\newblock Single-view {3D} reconstruction: A survey of deep learning methods.
\newblock \emph{{Computers \& Graphics}}, 94:\penalty0 164--190, 2021.

\bibitem[Furukawa \& Hern{\'a}ndez(2015)Furukawa and
  Hern{\'a}ndez]{mvs_tutorial}
Yasutaka Furukawa and Carlos Hern{\'a}ndez.
\newblock Multi-view stereo: A tutorial.
\newblock \emph{{Foundations and Trends{\textregistered} in Computer Graphics
  and Vision}}, 9\penalty0 (1-2):\penalty0 1--148, 2015.

\bibitem[Harley et~al.(2020)Harley, Lakshmikanth, Li, Zhou, Tung, and
  Fragkiadaki]{harley2019learning}
Adam~W Harley, Shrinidhi~K Lakshmikanth, Fangyu Li, Xian Zhou, Hsiao-Yu~Fish
  Tung, and Katerina Fragkiadaki.
\newblock Learning from unlabelled videos using contrastive predictive neural
  {3D} mapping.
\newblock \emph{{International Conference on Learning Representations}}, 2020.

\bibitem[He et~al.(2016)He, Zhang, Ren, and Sun]{resnets}
Kaiming He, Xiangyu Zhang, Shaoqing Ren, and Jian Sun.
\newblock Deep residual learning for image recognition.
\newblock In \emph{{Proceedings of the IEEE conference on Computer Vision and
  Pattern Recognition}}, pp.\  770--778, 2016.

\bibitem[He et~al.(2018)He, Gkioxari, Doll{\'a}r, and Girshick]{he2017mask}
Kaiming He, Georgia Gkioxari, Piotr Doll{\'a}r, and Ross Girshick.
\newblock Mask {R-CNN}.
\newblock \emph{{IEEE Transactions on Pattern Analysis and Machine
  Intelligence}}, 42\penalty0 (2):\penalty0 386--397, 2018.

\bibitem[He et~al.(2020)He, Fan, Wu, Xie, and Girshick]{moco}
Kaiming He, Haoqi Fan, Yuxin Wu, Saining Xie, and Ross Girshick.
\newblock Momentum contrast for unsupervised visual representation learning.
\newblock In \emph{Proceedings of the IEEE/CVF Conference on Computer Vision
  and Pattern Recognition}, pp.\  9729--9738, 2020.

\bibitem[He et~al.(2023)He, Chen, Cao, Yang, Cao, Li, Tang, Zhuang, and
  Lu]{srres_he2023single}
Zewei He, Du~Chen, Yanpeng Cao, Jiangxin Yang, Yanlong Cao, Xin Li, Siliang
  Tang, Yueting Zhuang, and Zhe-ming Lu.
\newblock Single image super-resolution based on progressive fusion of
  orientation-aware features.
\newblock \emph{{Pattern Recognition}}, 133:\penalty0 109038, 2023.

\bibitem[Hudson \& Manning(2018)Hudson and Manning]{mac}
Drew~A Hudson and Christopher~D Manning.
\newblock Compositional attention networks for machine reasoning.
\newblock In \emph{{International Conference on Learning Representations}},
  2018.

\bibitem[Jaderberg et~al.(2015)Jaderberg, Simonyan, Zisserman, et~al.]{stn}
Max Jaderberg, Karen Simonyan, Andrew Zisserman, et~al.
\newblock Spatial transformer networks.
\newblock In \emph{{Advances in Neural Information Processing Systems}}, pp.\
  2017--2025, 2015.

\bibitem[Jo et~al.(2015)Jo, Choi, Kim, and Kim]{srface_jo2015single}
Jaeik Jo, Heeseung Choi, Ig-Jae Kim, and Jaihie Kim.
\newblock Single-view-based {3D} facial reconstruction method robust against
  pose variations.
\newblock \emph{{Pattern Recognition}}, 48\penalty0 (1):\penalty0 73--85, 2015.

\bibitem[Johnson et~al.(2017)Johnson, Hariharan, van~der Maaten, Fei-Fei,
  Lawrence~Zitnick, and Girshick]{clevr}
Justin Johnson, Bharath Hariharan, Laurens van~der Maaten, Li~Fei-Fei,
  C~Lawrence~Zitnick, and Ross Girshick.
\newblock {CLEVR}: A diagnostic dataset for compositional language and
  elementary visual reasoning.
\newblock In \emph{{Proceedings of the IEEE Conference on Computer Vision and
  Pattern Recognition}}, pp.\  2901--2910, 2017.

\bibitem[Kamath et~al.(2021)Kamath, Singh, LeCun, Synnaeve, Misra, and
  Carion]{mdetr}
Aishwarya Kamath, Mannat Singh, Yann LeCun, Gabriel Synnaeve, Ishan Misra, and
  Nicolas Carion.
\newblock {MDETR}-modulated detection for end-to-end multi-modal understanding.
\newblock In \emph{{Proceedings of the IEEE/CVF International Conference on
  Computer Vision}}, pp.\  1780--1790, 2021.

\bibitem[Kang et~al.(2016)Kang, Yau, and Taylor]{srpose_kang2016simultaneous}
Xin Kang, Wai-Pan Yau, and Russell~H Taylor.
\newblock Simultaneous pose estimation and patient-specific model
  reconstruction from single image using maximum penalized likelihood
  estimation {(MPLE)}.
\newblock \emph{{Pattern Recognition}}, 57:\penalty0 61--69, 2016.

\bibitem[Kato et~al.(2018)Kato, Ushiku, and Harada]{neural_mesh_renderer}
Hiroharu Kato, Yoshitaka Ushiku, and Tatsuya Harada.
\newblock Neural {3D} mesh renderer.
\newblock In \emph{{Proceedings of the IEEE Conference on Computer Vision and
  Pattern Recognition}}, pp.\  3907--3916, 2018.

\bibitem[Kingma \& Ba(2015)Kingma and Ba]{adam}
Diederik~P Kingma and Jimmy Ba.
\newblock {ADAM}: A method for stochastic optimization.
\newblock In \emph{ICLR (Poster)}, 2015.

\bibitem[Kottur et~al.(2019)Kottur, Moura, Parikh, Batra, and
  Rohrbach]{clevr_dialog}
Satwik Kottur, Jos{\'e}~MF Moura, Devi Parikh, Dhruv Batra, and Marcus
  Rohrbach.
\newblock {CLEVR-Dialog}: A diagnostic dataset for multi-round reasoning in
  visual dialog.
\newblock In \emph{NAACL-HLT (1)}, 2019.

\bibitem[Liu et~al.(2018)Liu, Lehman, Molino, Petroski~Such, Frank, Sergeev,
  and Yosinski]{coordconv}
Rosanne Liu, Joel Lehman, Piero Molino, Felipe Petroski~Such, Eric Frank, Alex
  Sergeev, and Jason Yosinski.
\newblock An intriguing failing of convolutional neural networks and the
  {CoordConv} solution.
\newblock \emph{{Advances in Neural Information Processing Systems}}, 31, 2018.

\bibitem[Lombardi et~al.(2019)Lombardi, Simon, Saragih, Schwartz, Lehrmann, and
  Sheikh]{voxel_facebook}
Stephen Lombardi, Tomas Simon, Jason Saragih, Gabriel Schwartz, Andreas
  Lehrmann, and Yaser Sheikh.
\newblock Neural volumes: Learning dynamic renderable volumes from images.
\newblock \emph{{ACM Transactions on Graphics (TOG)}}, 38\penalty0
  (4):\penalty0 65, 2019.

\bibitem[Mildenhall et~al.(2021)Mildenhall, Srinivasan, Tancik, Barron,
  Ramamoorthi, and Ng]{nerf}
Ben Mildenhall, Pratul~P Srinivasan, Matthew Tancik, Jonathan~T Barron, Ravi
  Ramamoorthi, and Ren Ng.
\newblock {NERF}: Representing scenes as neural radiance fields for view
  synthesis.
\newblock \emph{Communications of the ACM}, 65\penalty0 (1):\penalty0 99--106,
  2021.

\bibitem[Nguyen-Phuoc et~al.(2019)Nguyen-Phuoc, Li, Theis, Richardt, and
  Yang]{hologan}
Thu Nguyen-Phuoc, Chuan Li, Lucas Theis, Christian Richardt, and Yong-Liang
  Yang.
\newblock {HoloGAN}: Unsupervised learning of {3D} representations from natural
  images.
\newblock In \emph{Proceedings of the IEEE/CVF International Conference on
  Computer Vision}, pp.\  7588--7597, 2019.

\bibitem[Nie et~al.(2020)Nie, Guo, Chang, Han, Huang, Hu, and
  Zhang]{nie2020shallow2deep}
Yinyu Nie, Shihui Guo, Jian Chang, Xiaoguang Han, Jiahui Huang, Shi-Min Hu, and
  Jian~Jun Zhang.
\newblock {Shallow2Deep}: Indoor scene modeling by single image understanding.
\newblock \emph{{Pattern Recognition}}, 103:\penalty0 107271, 2020.

\bibitem[Oord et~al.(2018)Oord, Li, and Vinyals]{oord2018representation}
Aaron van~den Oord, Yazhe Li, and Oriol Vinyals.
\newblock Representation learning with contrastive predictive coding.
\newblock \emph{arXiv preprint arXiv:1807.03748}, 2018.

\bibitem[Park et~al.(2019)Park, Darrell, and Rohrbach]{robust_caption}
Dong~Huk Park, Trevor Darrell, and Anna Rohrbach.
\newblock Robust change captioning.
\newblock In \emph{{Proceedings of the IEEE International Conference on
  Computer Vision}}, pp.\  4624--4633, 2019.

\bibitem[Perez et~al.(2018)Perez, Strub, De~Vries, Dumoulin, and
  Courville]{film}
Ethan Perez, Florian Strub, Harm De~Vries, Vincent Dumoulin, and Aaron
  Courville.
\newblock {FILM}: Visual reasoning with a general conditioning layer.
\newblock In \emph{{Proceedings of the AAAI Conference on Artificial
  Intelligence}}, volume~32, 2018.

\bibitem[Pontes et~al.(2018)Pontes, Kong, Sridharan, Lucey, Eriksson, and
  Fookes]{srmesh_pontes2018image2mesh}
Jhony~K Pontes, Chen Kong, Sridha Sridharan, Simon Lucey, Anders Eriksson, and
  Clinton Fookes.
\newblock Image2mesh: A learning framework for single image {3D}
  reconstruction.
\newblock In \emph{{Asian Conference on Computer Vision}}, pp.\  365--381.
  Springer, 2018.

\bibitem[Qi et~al.(2017)Qi, Su, Mo, and Guibas]{pointnet}
Charles~R Qi, Hao Su, Kaichun Mo, and Leonidas~J Guibas.
\newblock Pointnet: Deep learning on point sets for {3D} classification and
  segmentation.
\newblock In \emph{{Proceedings of the IEEE Conference on Computer Vision and
  Pattern Recognition}}, pp.\  652--660, 2017.

\bibitem[Qiu et~al.(2019)Qiu, Satoh, Suzuki, and Kataoka]{clevr_yue2019}
Yue Qiu, Yutaka Satoh, Ryota Suzuki, and Hirokatsu Kataoka.
\newblock Incorporating {3D} information into visual question answering.
\newblock In \emph{{2019 International Conference on 3D Vision (3DV)}}, pp.\
  756--765. IEEE, 2019.

\bibitem[Qiu et~al.(2020)Qiu, Satoh, Suzuki, Iwata, and Kataoka]{clevr_avs}
Yue Qiu, Yutaka Satoh, Ryota Suzuki, Kenji Iwata, and Hirokatsu Kataoka.
\newblock Multi-view visual question answering with active viewpoint selection.
\newblock \emph{Sensors}, 20\penalty0 (8):\penalty0 2281, 2020.

\bibitem[Rajeswar et~al.(2020)Rajeswar, Mannan, Golemo, Parent-L{\'e}vesque,
  Vazquez, Nowrouzezahrai, and Courville]{pix2shape}
Sai Rajeswar, Fahim Mannan, Florian Golemo, J{\'e}r{\^o}me Parent-L{\'e}vesque,
  David Vazquez, Derek Nowrouzezahrai, and Aaron Courville.
\newblock Pix2shape: Towards unsupervised learning of {3D} scenes from images
  using a view-based representation.
\newblock \emph{{International Journal of Computer Vision}}, pp.\  1--16, 2020.

\bibitem[Shepard \& Metzler(1971)Shepard and Metzler]{shepard1971mental}
Roger~N Shepard and Jacqueline Metzler.
\newblock Mental rotation of three-dimensional objects.
\newblock \emph{Science}, 171\penalty0 (3972):\penalty0 701--703, 1971.

\bibitem[Sitzmann et~al.(2019)Sitzmann, Thies, Heide, Nie{\ss}ner, Wetzstein,
  and Zollhofer]{deepvoxels}
Vincent Sitzmann, Justus Thies, Felix Heide, Matthias Nie{\ss}ner, Gordon
  Wetzstein, and Michael Zollhofer.
\newblock Deepvoxels: Learning persistent {3D} feature embeddings.
\newblock In \emph{{Proceedings of the IEEE/CVF Conference on Computer Vision
  and Pattern Recognition}}, pp.\  2437--2446, 2019.

\bibitem[Thies et~al.(2019)Thies, Zollh{\"o}fer, and
  Nie{\ss}ner]{deferred_neural}
Justus Thies, Michael Zollh{\"o}fer, and Matthias Nie{\ss}ner.
\newblock Deferred neural rendering: Image synthesis using neural textures.
\newblock \emph{{ACM Transactions on Graphics (TOG)}}, 38\penalty0
  (4):\penalty0 1--12, 2019.

\bibitem[Tian et~al.(2020)Tian, Krishnan, and Isola]{contrastive_multiview}
Yonglong Tian, Dilip Krishnan, and Phillip Isola.
\newblock Contrastive multiview coding.
\newblock In \emph{{European Conference on Computer Vision}}, pp.\  776--794.
  Springer, 2020.

\bibitem[Wang et~al.(2018)Wang, Zhang, Li, Fu, Liu, and Jiang]{pixel2mesh}
Nanyang Wang, Yinda Zhang, Zhuwen Li, Yanwei Fu, Wei Liu, and Yu-Gang Jiang.
\newblock {Pixel2Mesh}: Generating {3D} mesh models from single {RGB} images.
\newblock In \emph{{European Conference on Computer Vision}}, 2018.

\bibitem[Wu et~al.(2016)Wu, Zhang, Xue, Freeman, and Tenenbaum]{voxelgan}
Jiajun Wu, Chengkai Zhang, Tianfan Xue, Bill Freeman, and Josh Tenenbaum.
\newblock Learning a probabilistic latent space of object shapes via {3D}
  generative-adversarial modeling.
\newblock In \emph{{Advances in Neural Information Processing Systems}}, pp.\
  82--90, 2016.

\bibitem[Yan et~al.(2016)Yan, Yang, Yumer, Guo, and
  Lee]{srmesh_yan2016perspective}
Xinchen Yan, Jimei Yang, Ersin Yumer, Yijie Guo, and Honglak Lee.
\newblock Perspective transformer nets: Learning single-view {3D} object
  reconstruction without {3D} supervision.
\newblock \emph{{Advances in Neural Information Processing Systems}}, 29, 2016.

\bibitem[Yang et~al.(2022)Yang, Han, Zhang, and Tian]{yang2022exploring}
Yang Yang, Junwei Han, Dingwen Zhang, and Qi~Tian.
\newblock Exploring rich intermediate representations for reconstructing {3D}
  shapes from {2D} images.
\newblock \emph{Pattern Recognition}, 122:\penalty0 108295, 2022.

\bibitem[Yao et~al.(2018)Yao, Hsu, Zhu, Wu, Torralba, Freeman, and
  Tenenbaum]{inverse_gfx}
Shunyu Yao, Tzu~Ming Hsu, Jun-Yan Zhu, Jiajun Wu, Antonio Torralba, Bill
  Freeman, and Josh Tenenbaum.
\newblock {3D}-aware scene manipulation via inverse graphics.
\newblock \emph{{Advances in Neural Information Processing Systems}}, 31, 2018.

\bibitem[Yi et~al.(2018)Yi, Wu, Gan, Torralba, Kohli, and Tenenbaum]{nsvqa}
Kexin Yi, Jiajun Wu, Chuang Gan, Antonio Torralba, Pushmeet Kohli, and
  Joshua~B. Tenenbaum.
\newblock Neural-symbolic {VQA}: Disentangling reasoning from vision and
  language understanding.
\newblock In \emph{{Advances in Neural Information Processing Systems}}, pp.\
  1039--1050, 2018.

\bibitem[Yi et~al.(2019)Yi, Gan, Li, Kohli, Wu, Torralba, and
  Tenenbaum]{clevrer}
Kexin Yi, Chuang Gan, Yunzhu Li, Pushmeet Kohli, Jiajun Wu, Antonio Torralba,
  and Joshua~B Tenenbaum.
\newblock {CLEVRER:} collision events for video representation and reasoning.
\newblock In \emph{{International Conference on Learning Representations}},
  2019.

\end{thebibliography}
\bibliographystyle{iclr_style}


\newpage

\section{Supplementary material}

\subsection{Hyperparameters}

Table \ref{tb:hps} lists hyperparameters of the FILM module, which can be found in \texttt{architectures/clevr/probe.py}. \footnote{While an effort was made for this information to be accurate, the source code should always be the definitive reference.}

\begin{table}[]
    \centering
    \caption{Names and descriptions of hyperparameters used for the experiments. Please see Appendix Figure \ref{fig:hp_list} for more details.}
    \label{tb:hps}
    \begin{tabular}{m{0.3\linewidth}m{0.6\linewidth}}
    \toprule
    Hyperparameter & Description \\
    \midrule
    \texttt{rnn\_dim} & Output GRU embedding dimension \\[4pt]
    \texttt{rnn\_num\_layers} & Number of hidden layers in the GRU \\[4pt]
    \texttt{n\_resblocks} & How many FILMed ResBlocks do we use? \\[4pt]
    \texttt{with\_coords} & Do we append CoordConv \cite{coordconv} feature maps in each ResBlock? \\[4pt]
    \texttt{nf\_} & Number of output feature maps in each ResBlock \\[4pt]
    \texttt{with\_camera} & Do we concatenate the camera embedding with the GRU embedding? (This should be set to true when using camera-conditioned FILM.) \\[4pt]
    \texttt{ncf} & Dimension of the camera embedding. If set to none, we will simply use the original six coordinates rather than a projection MLP. \\[4pt]
    \texttt{weight\_decay} & Weight decay term for ADAM optimiser.\\[4pt]
    \texttt{imagenet\_scaling} & (For pretrained ImageNet encoder only) Use ImageNet mean/variance to scale the inputs instead of the default [-1, 1] scaling? \\[4pt]
    \bottomrule
    \end{tabular}
\end{table}

Each experiment was trained for a maximum of 60 epochs with the ADAM optimiser \citep{adam}, with a default learning rate of $3 \times 10^{-4}$ and first and second moment coefficients $\{\beta_{1}, \beta_{2}\} = \{0.9, 0.999\}$. Figure \ref{fig:hp_list} illustrates an example range of hyperparameters explored per experiment (where each experiment refers to one of the results performed in Table \ref{tb:results}). Note that this is \emph{not} an exhaustive range of values explored -- rather, the values seen per experiment in the Figure \ref{fig:hp_list} correspond to the batch of runs in which at least one of the runs inside the batch gave the highest validation score(s). (Other batches of HP/value combinations were also run but may not have yielded the best validation accuracies, and therefore are not shown in the figure.) Whichever experiment was found to have the highest validation score was re-trained multiple times (under different seeds) and evaluated on the test set.

\begin{figure}[h]
\begin{lstlisting}[language={},  
  basicstyle=\tiny\ttfamily,
  columns=fullflexible,
  frame=single,
  breaklines=true,
  postbreak=\mbox{\textcolor{red}{$\hookrightarrow$}\space}]
2d film, no camera
------------------

  {'rnn_dim': {128, 512, 256, 1024}, 'n_resblocks': {4}, 'encoder': {'gru'}, 'with_coords': {True}, 'rnn_num_layers': {1, 2}, 'n_in': {1024}, 'nf': {None}, 'coord_shape': {(14, 14)}, 'with_camera': {False}, 'ncf': {None}}
  {'weight_decay': {1e-05, 0.0001}}

2d film, camera
---------------

  {'rnn_dim': {1024, 512}, 'n_resblocks': {4}, 'encoder': {'gru'}, 'with_coords': {True}, 'rnn_num_layers': {1, 2}, 'n_in': {1024}, 'nf': {64, 128}, 'coord_shape': {(14, 14)}, 'with_camera': {True}, 'ncf': {64}}
  {'weight_decay': {0, 0.0001, 1e-05}}
  {'imagenet_scaling': {False}}
  {'batch_size': {16}}
  

3d film, no camera embed/rotation
---------------------------------

  {'rnn_dim': {512, 1024}, 'rnn_num_layers': {1}, 'n_resblocks': {4}, 'encoder': {'gru'}, 'with_coords': {True}, 'coord_shape': {(16, 14, 14)}, 'with_camera': {False}, 'ncf': {None}, 'flatten_3d': {False}, 'is_3d': {True}, 'n_in': {128}, 'nf': {64, 128}}
  {'weight_decay': {0, 1e-05}}
  {'imagenet_scaling': {True}}
  
3d film, camera rotation
------------------------

  {'rnn_dim': {512, 1024}, 'rnn_num_layers': {1, 2}, 'n_resblocks': {4}, 'encoder': {'gru'}, 'with_coords': {True}, 'coord_shape': {(16, 14, 14)}, 'with_camera': {False}, 'ncf': {None}, 'flatten_3d': {False}, 'is_3d': {True}, 'n_in': {128}, 'nf': {64, 128}}
  {'weight_decay': {0, 0.0001, 1e-05}}
  {'imagenet_scaling': {True}}

3d film, camera embed
---------------------

  {'rnn_dim': {512, 1024}, 'rnn_num_layers': {1, 2}, 'n_resblocks': {4}, 'encoder': {'gru'}, 'with_coords': {True}, 'coord_shape': {(16, 14, 14)}, 'with_camera': {True}, 'ncf': {None}, 'flatten_3d': {False}, 'is_3d': {True}, 'n_in': {128}, 'nf': {64, 128}}
  {'weight_decay': {1e-05, 0, 1e-06}}
  {'imagenet_scaling': {True}}


3d film, both camera embed/rotation
-----------------------------------

  {'rnn_dim': {512, 1024}, 'rnn_num_layers': {1}, 'n_resblocks': {4}, 'encoder': {'gru'}, 'with_coords': {True}, 'coord_shape': {(16, 14, 14)}, 'with_camera': {True}, 'ncf': {64}, 'flatten_3d': {False}, 'is_3d': {True}, 'n_in': {128}, 'nf': {64, 128}}
  {'weight_decay': {0, 1e-05, 1e-06}}
  {'imagenet_scaling': {True}}
\end{lstlisting}
\caption{The range of hyperparameters explored per experiment in Table \ref{tb:results} (the upper bound canonical baseline and contrastive experiments are not shown here).}
\label{fig:hp_list}
\end{figure}

\subsection{MAC baselines} \label{supp:mac}

Code for our MAC baseline was adapted from here\footnote{\url{https://github.com/rosinality/mac-network-pytorch.git}}. In short, a bidirectional LSTM was used here with 12 time steps used for the MAC reasoning step. In order to leverage camera information, we simply concatenated the camera's embedding to the summary question embedding (not to the contextual word embeddings). For the best performing experiment, self-attention was disabled.

Due to significant time constraints, hyperparameter tuning for this baseline was minimal. However, we can see in Table \ref{tb:results} that it obtains similar validation accuracy to the 2D FILM + camera conditioning baseline, and in principle can probably also be adapted to perform 3D reasoning.

\subsection{Example images from dataset}

\begin{figure}[ht]
    \centering
    \includegraphics[width=0.18\textwidth]{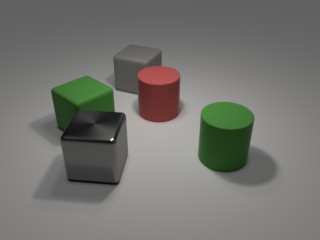}
    \includegraphics[width=0.18\textwidth]{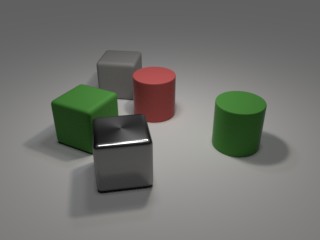}
    \includegraphics[width=0.18\textwidth]{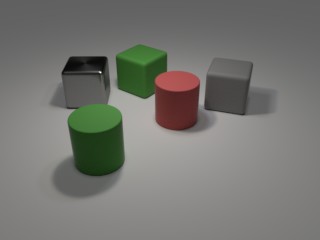}
    \includegraphics[width=0.18\textwidth]{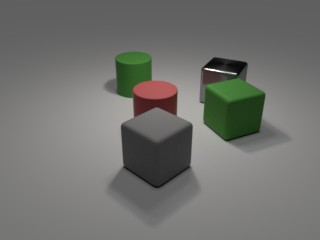}
    \includegraphics[width=0.18\textwidth]{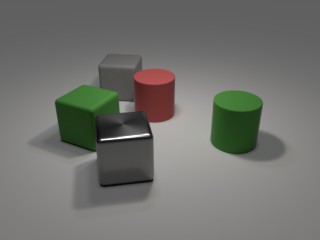} \\ \vspace{0.2cm}
    \includegraphics[width=0.18\textwidth]{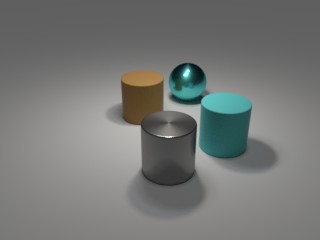}
    \includegraphics[width=0.18\textwidth]{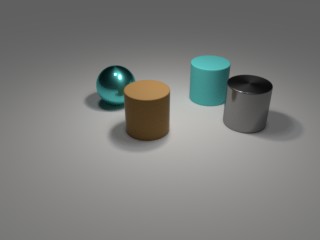}
    \includegraphics[width=0.18\textwidth]{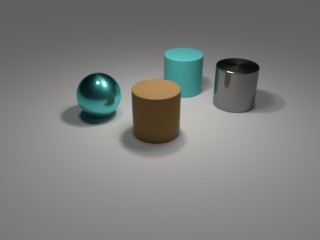}
    \includegraphics[width=0.18\textwidth]{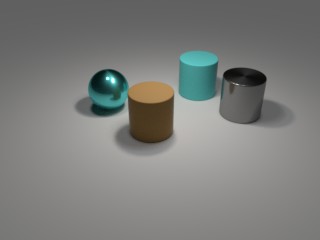}
    \includegraphics[width=0.18\textwidth]{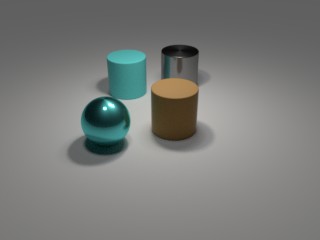}
    \caption{Random views of an example scene in CLEVR-MRT. The center image is the `canonical' view.}
    \label{fig:clevr_kiwi_more_examples}
\end{figure}

See Figure \ref{fig:clevr_kiwi_more_examples}.

\newcommand{\exampleone}{figures/clevr_kiwi/examples_098}
\newcommand{\exampletwo}{figures/clevr_kiwi/examples_900}

\end{document}